\documentclass[sigconf]{acmart}
\usepackage{capt-of}
\usepackage{algorithm}
\usepackage{algorithmic}
\usepackage{graphicx}
\usepackage{amsmath}
\usepackage{amsthm} 
\usepackage{color}
\usepackage{bm}
\newtheorem{theorem}{Theorem}
\newtheorem{definition}{Definition}
\usepackage{tcolorbox}
\usepackage{url}

\makeatletter
\DeclareFontFamily{OMX}{MnSymbolE}{}
\DeclareSymbolFont{MnLargeSymbols}{OMX}{MnSymbolE}{m}{n}
\SetSymbolFont{MnLargeSymbols}{bold}{OMX}{MnSymbolE}{b}{n}
\DeclareFontShape{OMX}{MnSymbolE}{m}{n}{
    <-6>  MnSymbolE5
   <6-7>  MnSymbolE6
   <7-8>  MnSymbolE7
   <8-9>  MnSymbolE8
   <9-10> MnSymbolE9
  <10-12> MnSymbolE10
  <12->   MnSymbolE12
}{}
\DeclareFontShape{OMX}{MnSymbolE}{b}{n}{
    <-6>  MnSymbolE-Bold5
   <6-7>  MnSymbolE-Bold6
   <7-8>  MnSymbolE-Bold7
   <8-9>  MnSymbolE-Bold8
   <9-10> MnSymbolE-Bold9
  <10-12> MnSymbolE-Bold10
  <12->   MnSymbolE-Bold12
}{}

\let\llangle\@undefined
\let\rrangle\@undefined
\DeclareMathDelimiter{\llangle}{\mathopen}%
                     {MnLargeSymbols}{'164}{MnLargeSymbols}{'164}
\DeclareMathDelimiter{\rrangle}{\mathclose}%
                     {MnLargeSymbols}{'171}{MnLargeSymbols}{'171}
\makeatother

\usepackage{multirow}

\DeclareMathOperator{\sign}{sign}

\newcommand{\sfI}{\mathsf{I}}

\newcommand{\bme}{\bm{e}}

\newcommand{\bmg}{\bm{g}}

\newcommand{\bmr}{\bm{r}}

\newcommand{\bmu}{\bm{u}}

\newcommand{\bmx}{\bm{x}}

\newcommand{\rmc}{\mathrm{c}}
\newcommand{\rmd}{\mathrm{d}}

\newcommand{\bmzero}{\bm{0}}

\newcommand{\bmbeta}{\bm{\beta}}

\newcommand{\bmdelta}{\bm{\delta}}

\newcommand{\bmxi}{\bm{\xi}}

\newcommand{\calD}{\mathcal{D}}

\newcommand{\calN}{\mathcal{N}}

\newcommand{\calS}{\mathcal{S}}

\AtBeginDocument{%
  \providecommand\BibTeX{{%
    \normalfont B\kern-0.5em{\scshape i\kern-0.25em b}\kern-0.8em\TeX}}}

\setcopyright{acmcopyright}
\copyrightyear{2023} 
\acmYear{2023} 
\setcopyright{rightsretained} 
\acmConference[KDD '23]{Proceedings of the 29th ACM SIGKDD Conference on Knowledge Discovery and Data Mining}{August 6--10, 2023}{Long Beach, CA, USA}
\acmBooktitle{Proceedings of the 29th ACM SIGKDD Conference on Knowledge Discovery and Data Mining (KDD '23), August 6--10, 2023, Long Beach, CA, USA}
\acmDOI{10.1145/3580305.3599365}
\acmISBN{979-8-4007-0103-0/23/08}

\makeatletter
\gdef\@copyrightpermission{
  \begin{minipage}{0.3\columnwidth}
    \href{https://creativecommons.org/licenses/by/4.0/}
    {\includegraphics[width=0.90\textwidth]{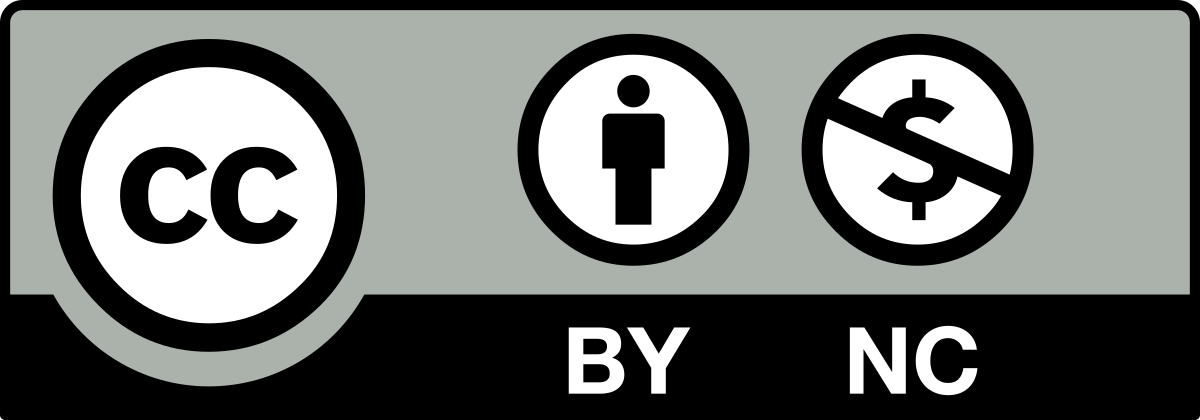}}
  \end{minipage}\hfill
  \begin{minipage}{0.7\columnwidth}
    \href{https://creativecommons.org/licenses/by/4.0/}{This work is licensed under a Creative Commons Attribution 4.0 International License.}
  \end{minipage}
  \vspace{5pt}
}
\makeatother


\begin{document}

\title{Generative Perturbation Analysis for Probabilistic Black-Box Anomaly Attribution}

\author{Tsuyoshi Id\'e}
\email{tide@us.ibm.com}
\orcid{0000-0001-8993-2776}
\affiliation{%
  \institution{IBM Research, Thomas J. Watson Research Center}
  \streetaddress{}
  \city{Yorktown Heights}
  \state{New York}
  \country{USA}
  \postcode{10598}
}

\author{Naoki Abe}
\email{nabe@us.ibm.com}
\orcid{0000-0002-4048-3989}
\affiliation{%
  \institution{IBM Research, Thomas J. Watson Research Center}
  \streetaddress{}
  \city{Yorktown Heights}
  \state{New York}
  \country{USA}
  \postcode{10598}
}


\begin{abstract}
We address the task of probabilistic anomaly attribution in the black-box regression setting, where the goal is to compute the probability distribution of the attribution score of each input variable, given an observed anomaly. The training dataset is assumed to be unavailable. This task differs from the standard XAI (explainable AI) scenario, since we wish to explain the anomalous deviation from a black-box prediction rather than the black-box model itself. 

We begin by showing that mainstream model-agnostic explanation methods, such as the Shapley values, are not suitable for this task because of their ``deviation-agnostic property.'' 
We then propose a novel framework for probabilistic anomaly attribution that allows us to not only compute attribution scores as the predictive mean but also quantify the uncertainty of those scores. This is done by considering a generative process for perturbations that counter-factually bring the observed anomalous observation back to normalcy. We introduce a variational Bayes algorithm for deriving the distributions of per variable attribution scores. To the best of our knowledge, this is the first probabilistic anomaly attribution framework that is free from being deviation-agnostic. 
\end{abstract}


\maketitle

\section{Introduction}\label{sec:Introduction}

Over the last decade, we have witnessed a dramatic resurgence of deep neural networks (DNNs) and numerous attempts to use DNNs in real-world applications. Despite their remarkable achievements, growing concerns are also expressed regarding the lack of transparency in advanced machine learning (ML) algorithms, making explainable artificial intelligence (XAI) an active research area in the data mining community. While early XAI studies tended to focus on the psychological aspects of how AI should be made explainable, the bulk of research interest is now shifting towards {\em actionability} in business and industrial applications, as the adoption of AI is becoming more widespread~\cite{dw2019darpa,langer2021we}. 

\begin{figure}[t]
\begin{center}
\includegraphics[trim={0.9cm 4.cm 0.5cm 1cm},clip,width=8.5cm]{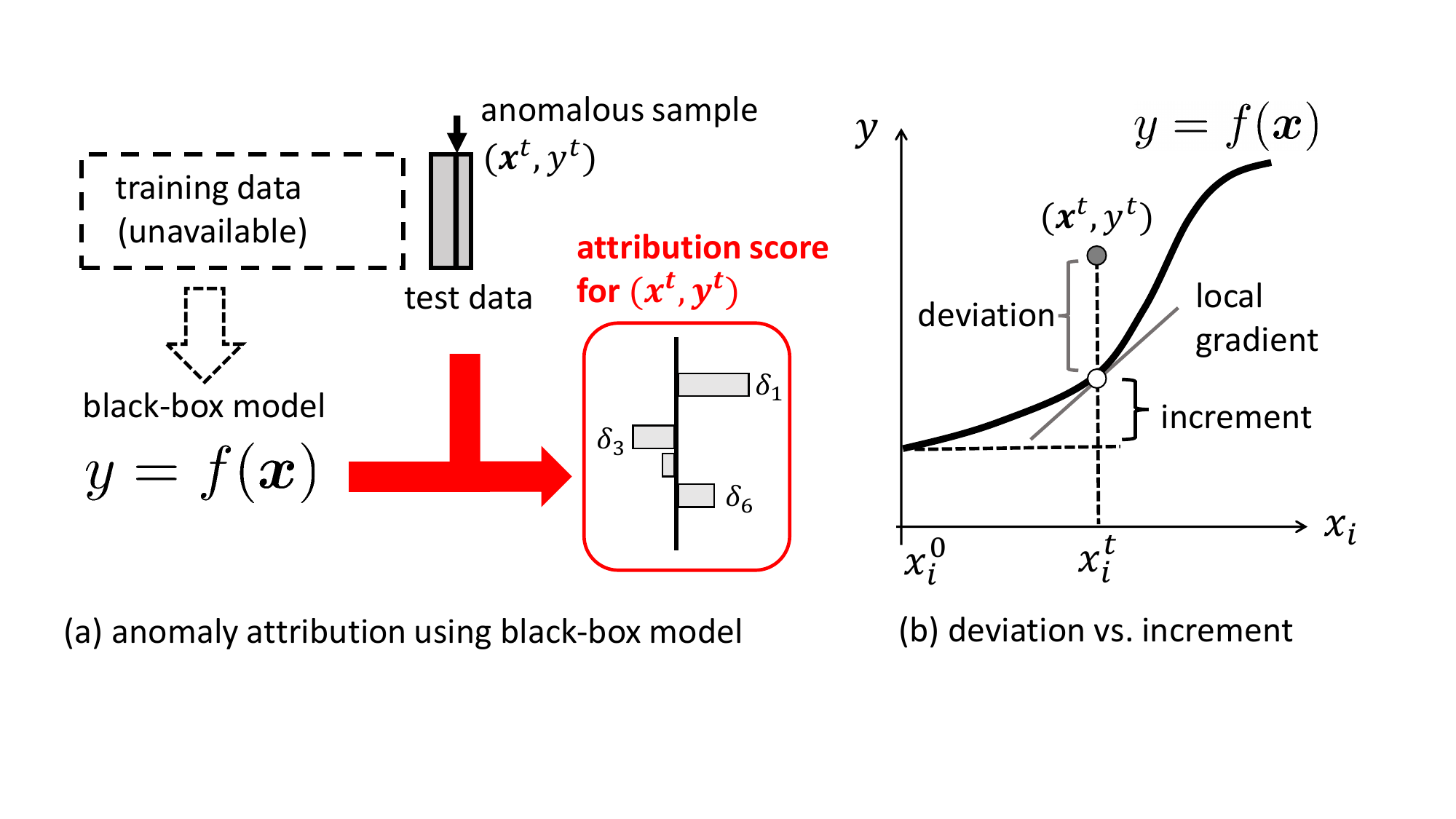}
\end{center}
\caption{Problem setting and motivation. (a) Given a black-box deterministic regression model and anomalous sample(s), our goal is to find the \textit{probability distribution} of input variables' responsibility scores without access to the training data. (b)~Existing attribution methods attempt to explain either the local gradient or the increment from a reference point $\bmx^0$, rather than the deviation of the sample in question.}
\label{fig:problem_setting_and_deviation_v_increment}
\vspace{-0.2cm}
\end{figure}

One important problem in this context is how to explain an unusual event, observed as a significant discrepancy from the prediction of an ML model. Although this problem encompasses various different scenarios, we are particularly interested in the task of \textbf{anomaly attribution in the doubly black-box regression setting} (see Fig.~\ref{fig:problem_setting_and_deviation_v_increment}~(a)): We are given a black-box regression model $y=f(\bmx)$, where $y$ is the real-valued noisy output (such as miles per gallon) and $\bmx$ is a vector of noisy real-valued input variables (such as driver's weight and average speed). We have access to the API (application programming interface) of $f(\cdot)$ but do not have access to either its parametric form or training data (hence, ``doubly''). Given a limited amount of test samples, we ask: how can we quantify the contribution of each input variable in the face of an unexpected deviation between observation and prediction?

This question has typically been addressed with one of the following three model-agnostic post-hoc XAI methods in the literature: 
1) Local linear surrogate modeling, which is best known under the name LIME (Local Interpretable Model-agnostic Explanations)~\cite{ribeiro2016should}; 
2) Shapley value (SV), which was first introduced to the ML community by~\cite{kononenko2010efficient}; and
3) integrated gradient (IG)~\cite{sundararajan2017axiomatic}. 

Despite their popularity, however, there are two major limitations with those methods. One is that all of them are, in fact, ``\textbf{deviation-agnostic},'' meaning that they explain the black-box function $f(\cdot)$ itself in the form of the local gradient or an increment, not the observed deviation, as illustrated in Fig.~\ref{fig:problem_setting_and_deviation_v_increment}~(b). Here, note that, unlike the standard XAI scenarios, we seek explanations \textit{relative to} the deviation from a black-box prediction, as we will discuss in detail later. The other limitation is that they have limited capabilities of quantifying the uncertainty of the attribution scores. Motivated by the requirements from industrial applications, e.g.,~\cite{onchis2020should}, uncertainty quantification (UQ) of attribution scores is becoming a major topic in XAI research. In the black-box setting without access to the training data, however, this problem is considered extremely challenging and limited work has been done to date. Existing works include empirical comparative studies, e.g.,~\cite{zhang2019should,zhou2022feature}, and semi-theoretical analysis based on known results of probabilistic linear regression~\cite{guo2018explaining,Garreau20AISTATS,zhao2021baylime,slack2021reliable}.

In this paper, we propose a novel probabilistic framework called the \textbf{generative perturbation analysis} (GPA) for anomaly attribution in the black-box regression setting, which we believe is the first fully probabilistic black-box attribution algorithm. The key idea is to consider a counterfactual data generative process including perturbation $\bmdelta$ as a model parameter, and reduce the task of attribution to that of statistical parameter estimation. In this way, the uncertainty in attribution is naturally evaluated by finding its posterior distribution. Here we additionally introduce a novel idea of using variational Bayes inference to decompose the contribution of each of the input variables.

To summarize, our contributions are: 1) to mathematically show that the existing attribution methods have the deviation-agnostic property; 2) to uncover their interrelationship that has been hitherto unnoticed; and 3) to propose the first generative framework for anomaly attribution.

\begin{table*}[tb]
    \centering
    \caption{Comparison of model-agnostic attribution methods in the regression setting. }
    \vspace{-0.3cm}
    \small
    \begin{tabular}{c c c c c c c}
    \hline \hline
            & model-agnostic & training-data-free & baseline-input-free & $y$-sensitive & built-in UQ & reference point \\
    \hline
     LIME~\cite{ribeiro2016should}  & yes & yes    & yes    &no & yes/no & infinitesimal vicinity \\
     SV~\cite{kononenko2010efficient,vstrumbelj2014explaining}   & yes& no    & yes    &no &no & globally distributional\\ 
     IG~\cite{sundararajan2017axiomatic,SippleICML2020}    & yes& yes    & no    &no &no & arbitrary\\
     EIG~\cite{deng2021unified}   & yes& no    & yes    &no &no & globally distributional \\
     Z-score~\cite{Chandola09AnomalySurvey} & yes& no    & yes    &no &no & global mean of predictors \\
    LC~\cite{Ide21AAAI} & yes& yes    & yes    &yes &no & maximum likelihood point  \\
     \textbf{GPA} & \textbf{yes}    & \textbf{yes}    & \textbf{yes}   &\textbf{yes} &\textbf{yes} & maximum a posteriori point \\
    \hline
    \end{tabular}
    \label{table:comparison chart}
\end{table*}

\section{Related Work}

Anomaly attribution has been studied as a sub-task of anomaly detection in the ML community, typically in the white-box unsupervised setting. In the supervised setting, the majority of prior works are about either model- or classification-specific algorithms. For example, saliency maps~\cite{simonyan2013deep,Selvaraju_2017_ICCV} and layer-wise relevance propagation~\cite{montavon2019layer} are well-known \textit{model-specific} attribution methods. Sainyam et al.~\cite{galhotra2021explaining} leveraged a counterfactual  framework~\cite{guidotti2022counterfactual} for probabilistic black-box explanations in the classification setting with binary variables. Similar approaches have been discussed under the terms like perturbation-based or mask-based (e.g.~\cite{fong2017interpretable,Petsiuk2018rise,fong2019understanding}), but most of them are for classification without the capability of computing the distribution of attribution score and are not directly applicable to the present setting. 

In the \textit{model-agnostic} regression setting, 1) local linear modeling, 2) SV, and 3) IG have been widely used for black-box attribution, as summarized in Table~\ref{table:comparison chart}, along with three additional methods: The expected integrated gradient (EIG)~\cite{deng2021unified}, which is a generalized version of IG, the $Z$-score, which is a standard outlier detection metric in the \textit{unsupervised} setting, and likelihood compensation (LC)~\cite{Ide21AAAI}, which conducts a semi-probabilistic analysis for attribution.  
In the context of \textit{anomaly} attribution, LIME and its variant have been applied to anomaly explanation~\cite{giurgiu2019additive,zhang2019ace}. SV is used in sensor fault diagnosis~\cite{hwang2021sfd} and for explaining unexpected observations in crop yield analysis~\cite{mariadass2022extreme} and unusual warranty claims~\cite{antwarg2021explaining}. Also, the use of IG for anomaly explanation is discussed by  Sipple~\cite{SippleICML2020,sipple2022general}.

Interestingly, it has been suggested that these attribution methods may have some mutual connection. Prior work along this line includes Deng et al.~\cite{deng2021unified}, which attempted to characterize IG using Taylor expansion and gave the first definition of EIG. Also, Sundararajan and Najmi~\cite{sundararajan2020many} proposed a unified attribution framework, where they pointed out that there can be a few different definitions for SV and discussed the relationship with IG in a qualitative manner. Lundberg and Lee~\cite{lundberg2017unified} reintroduced the SV-based attribution method~\cite{kononenko2010efficient} to propose a hybrid method between SV and LIME. Inspired by these works, we go one step further in this paper: We explicitly show a mathematical relationship between those existing attribution methods, and show that the deviation-agnostic property (see the `$y$-sensitive' column in Table~\ref{table:comparison chart}) is an inherent consequence of the common mathematical structure.

Another important contribution of this paper is the proposal of a principled framework for probabilistic prediction of attribution scores. Most of the existing works tackling this problem~\cite{Garreau20AISTATS,zhao2021baylime,slack2021reliable} under settings similar to ours use the standard result of probabilistic linear regression (see, e.g.,~Chap.~3 of~\cite{Bishop}) to evaluate uncertainty in the regression coefficients as the LIME attribution score (the 'built-in UQ' column in Table~\ref{table:comparison chart}). However, the black-box model $f(\bmx)$ is generally highly nonlinear; It is not clear to what extent the theoretical results of the linear model apply. Also, it is not clear how the distribution of the attribution score is computed for each input variable (hence `yes/no' in the table). In fact, BayLIME~\cite{zhao2021baylime}'s posterior covariance is a constant that depends only on the hyper-parameters independently of $f(\cdot)$ (See Sec.~\ref{sec:Distribution_analysis}). LC~\cite{Ide21AAAI} shares a similar starting point with ours but differs fundamentally in that it is not able to compute the probability distribution of the attribution score. Guo et al.~\cite{guo2018explaining} used a Dirichlet-enhanced probabilistic linear regression mixture but it is intended for global model explanations rather than local anomaly attribution.

\section{Problem Setting}\label{sec:problem_setting}

As mentioned earlier, we focus on the task of anomaly attribution in the \textit{regression} setting rather than classification or unsupervised settings. Figure~\ref{fig:problem_setting_and_deviation_v_increment}~(a) summarizes the overall problem setting. Suppose we have a (deterministic) regression model $y=f(\bm{x})$ in the {\em doubly black-box setting}: Neither the training data set $\calD_{\mathrm{train}}$ nor the (true) distribution of $\bmx$ is available (see the `training-data-free' column in Table~\ref{table:comparison chart}).  Throughout the paper, the input variable $\bm{x} \in \mathbb{R}^M$ and the output variable $y \in \mathbb{R}$ are assumed to be \textit{noisy real-valued}, where $M$ is the dimensionality of the input vector. We also assume that queries to get the response $f(\bm{x})$ can be performed cheaply at any~$\bm{x}$.

In practice, anomaly attribution is typically coupled with anomaly detection: When we observe a test sample $(\bmx,y)=(\bmx^t,y^t)$, we first compute an anomaly score $a^t = a(\bmx^t,y^t)$ to quantify how anomalous it is. Then, if $a^t \in \mathbb{R}$ is high enough, we go to the next step of anomaly attribution. In this scenario, the task of anomaly attribution is defined as follows. 
\begin{definition}[\textbf{probabilistic anomaly attribution}]
 Given a black-box regression model $y=f(\bm{x})$ and observed test sample(s), compute the distribution of the score for each input variable indicative of the extent to which that variable is responsible for the sample being anomalous.  
\end{definition}
We can readily generalize the problem to that of  \textit{collective} probabilistic anomaly detection and attribution. Specifically, given a test data set  $\mathcal{D}_\mathrm{test}= \{ (\bm{x}^t,y^t) \mid t=1,\ldots,N_\mathrm{test}\}$, where $t$ is the index for the $t$-th test sample and $N_\mathrm{test}$ is the number of test samples, we can consider anomaly score as well as attribution score distributions for the whole test set $\calD_{\mathrm{test}}$. 

The standard approach to \textbf{anomaly detection} is to use the negative log-likelihood of the test sample(s) as the anomaly score (See, e.g.,~\cite{lee2000information,Yamanishi2000,staniford2002practical,noto2010anomaly,yamanishi2004line}). Assume that, from the deterministic regression model, we can somehow obtain $p(y \mid \bm{x})$, a probability density over $y$ given the input signal $\bm{x}$. Under the i.i.d.~assumption, the anomaly score can be written as
\begin{gather}\label{eq:outlier_score}
\textstyle
a(\bmx^t,y^t) =  -\ln p(y^t \mid \bm{x}^t), \quad \mbox{or, } 
\\ \label{eq:changeScore}
\textstyle
a(\mathcal{D}_\mathrm{test}) =-
\frac{1}{
	N_\mathrm{test}
}\sum_{t=1}^{N_\mathrm{test}}\ln p(y^t \mid \bm{x}^t) ,
\end{gather}
corresponding to the single sample and collective cases, respectively. \textbf{Anomaly attribution} is the task to attribute a high anomaly score to each of the input variables.

\paragraph{Notation}
We use boldface to denote vectors. The $i$-th dimension of a vector $\bm{\delta}$ is denoted as $\delta_i$. The $\ell_1$ and $\ell_2$ norms of a vector are denoted by $\| \cdot \|_1$ and $\| \cdot \|_2$, respectively, and are defined as $\| \bm{\delta} \|_1 \triangleq \sum_i | \delta_i|$ and $\| \bm{\delta} \|_2 \triangleq \sqrt{\sum_i  \delta_i^2}$. The sign function $\mathrm{sign}(\delta_i) $ is defined as being $1$ for $\delta_i>0$, and $-1$ for $\delta_i <0$.
For $\delta_i=0$, the function takes an indeterminate value in $[-1,1]$. For a vector input, the definition applies element-wise, yielding a vector of the same size as the input vector.
We distinguish between a random variable and its realizations via the absence or presence of a superscript. For notational simplicity, we use $p(\cdot)$ or $P(\cdot)$ as a proxy to represent different probability distributions, whenever there is no confusion. For instance, $p(\bm{x})$ is used to represent the probability density of a random variable $\bm{x}$ while $p(y \mid \bm{x})$ is a different distribution of another random variable $y$ conditioned on $\bm{x}$. 

\section{Existing attribution methods are deviation-agnostic}
\label{sec:EIG}

This section summarizes our remarkable new results on the existing attribution methods: 1) IG, SV, and LIME are inherently deviation-agnostic and are not appropriate for anomaly attribution, 2) SV is equivalent to EIG up to the second order in the power expansion, and 3) LIME can be derived as the derivative of IG or EIG in a certain limit. Throughout this subsection, we assume that the derivative of the black-box regression function $f(\cdot)$ is computable somehow to an arbitrary order.

Formally, the deviation-agnostic property is defined as follows: 
\begin{definition}[\textbf{deviation-agnostic}]
An anomaly attribution method $A$ is said to be {\em deviation-agnostic} if for any black-box regression model $f(\cdot)$, observed test sample $(\bmx^t, y^t)$, deviation $\Delta$ and input variable $i$, 
$
A_{f,i}(\bmx^t, y^t) = A_{f,i}(\bmx^t, y^t+\Delta)
$,  
where $A_{f,i}(\bmx^t, y^t)$ denotes the attribution score computed by $A$ for $f$, $i$ and $(\bmx^t, y^t)$. 
\end{definition}
We often drop the subscript $f$ when it is clear from the context.

\subsection{Deviation-agnostic properties}\label{subsec:deviation_agnosticness}

\subsubsection{LIME}\label{subsubsec:LIME}
In general, the local linear surrogate modeling approach fits a linear regression model locally to explain a black-box function in the vicinity of a given test sample $(\bm{x}^t,y^t)$. For anomaly attribution, we need to consider the \textit{deviation function} $F(\bmx,y) \triangleq f(\bmx)-y$ instead of $f(\bmx)$. Algorithm~\ref{algo:LIME} summarizes the local anomaly attribution procedure. Let $\beta_i$ denote the $i$-th output by $\mathrm{LIME}_i(\bmx^t,y^t)$. Rather unexpectedly, despite the modification to fit $F(\bmx,y)$ rather than $f(\bmx)$, the following property holds:
\begin{theorem}\label{th:LIME_deviation_agnositic_main_text} LIME is deviation-agnostic: $\mathrm{LIME}_i(\bmx^t,y^t)=\mathrm{LIME}_i(\bmx^t)$. 
\end{theorem}
\begin{proof}
     With $\nu$ being the $\ell_1$ regularization strength, the loss function for LIME is written as
\begin{align*}
\Psi&(\bmbeta,\beta_0) = \frac{1}{N_s}\sum_{n=1}^{N_s} (z^{t[n]} - \beta_0 - \bmbeta^\top \bm{x}^{t[n]})^2 + \nu\| \bmbeta\|_1,
\\
&= \frac{1}{N_s}\sum_{n=1}^{N_s} (f(\bm{x}^{t[n]}) - (y^t+\beta_0) - \bmbeta^\top \bm{x}^{t[n]})^2 + \nu\| \bmbeta\|_1,
\end{align*}
which is equivalent to the lasso objective for LIME with the intercept $y^t + \beta_0$. Since the lasso objective is convex, the solution $\bmbeta$ is unique. With an arbitrary adjusted intercept, the attribution score $\bmbeta$ remains unchanged. Hence, $\forall i$, $\mathrm{LIME}_i(\bmx^t,y^t)=\mathrm{LIME}_i(\bmx^t)$. 
\end{proof}

In the local linear surrogate modeling approach, the final attribution score can vary depending on the nature of the regularization term. For the theoretical analysis below, we use a generic algorithm by setting $\nu \to 0_+$ in Algorithm~\ref{algo:LIME}, and call the resulting attribution score $\mathrm{LIME}^0_i$ for $i=1,\ldots,M$. As is well-known, $\mathrm{LIME}^0_i$ is a local estimator of $\partial f/\partial x_i$ at $\bmx=\bmx^t$ if $f$ is locally differentiable.

\begin{algorithm}[htb]
\caption{Local linear surrogate model for $F(\bmx,y)$}\label{algo:LIME}
\label{alg:LIME}
\begin{algorithmic}[1] 
\REQUIRE $f(\bm{x})$, test point $(\bm{x}^t,y^t)$,  regularization parameter $\nu$.
\STATE Randomly populate $N_s$ points $\{ \bm{x}^{t[1]}, \ldots, \bm{x}^{t[N_s]}\}$ in the vicinity of $\bm{x}^t$ ($N_s\sim 1000)$.
\STATE Compute the deviation $z^{t[n]} \triangleq f(\bm{x}^{t[n]}) -y^t$ for all $n$.
\STATE Fit a linear model $z = \beta_0 + {\bmbeta}^\top \bm{x}$ using the $\ell_1$ weight $\nu$ to the dataset $\{ (\bm{x}^{t[n]} , z^{t[n]}) \mid n=1,\ldots,N_s \}$.
\RETURN  ${\bmbeta}$, which is the local attribution score at $(\bm{x}^t,y^t)$.
\end{algorithmic}
\end{algorithm}

\subsubsection{Integrated gradient} \label{subsubsec:IG}

For anomaly attribution, which is an \textit{input} attribution task,  IG~\cite{sundararajan2017axiomatic,SippleICML2020} should be computed for the deviation function $F(\bmx,y) \triangleq f(\bmx)-y$ rather than $f$ alone as 
\begin{align}\label{eq:IG_x_y_main_text}
    \mathrm{IG}_i(\bmx^t,y^t \mid \bmx^0,y^0) 
    &\triangleq (x_i^t-x_i^0) \int_0^1\mathrm{d}\alpha  \left.\frac{\partial F}{\partial x_i}\right|_\alpha
\end{align}
for $i=1,\ldots,M$, where the gradient is estimated at $\bmx = \bmx^0 +(\bmx^t-\bmx^0)\alpha$ and $y =y^0+(y^t-y^0)\alpha$. The baseline input $(\bmx^0,y^0)$ has to be determined from prior knowledge. We also define EIG by integrating out the baseline input:
\begin{align}\label{eq:EIG_x_y_main_text}
    \mathrm{EIG}_i(\bmx^t,y^t) \!
    \triangleq \!\!
    \int\!\!\mathrm{d}y^0\!\!\!\int\!\!\mathrm{d}\bmx^0
    P(\bmx^0,y^0)\mathrm{IG}_i(\bmx^t,y^t \mid \bmx^0,y^0), 
\end{align}
where $P(\bmx,y)$ is the joint distribution of $\bmx$ and $y$, which is actually unavailable in our setting. The following property holds:
\begin{theorem} \label{th:IG_EIG_deviation_agnositic_main_text}
IG and EIG are deviation-agnostic: $\mathrm{IG}_i(\bmx^t,y^t) = \mathrm{IG}_i(\bmx^t)$ and $\mathrm{EIG}_i(\bmx^t,y^t) = \mathrm{EIG}_i(\bmx^t)$.
\end{theorem}
\begin{proof}
We define 
\begin{align}\label{eq:IG-def_main_text}
    \mathrm{IG}_i(\bmx^t \mid \bmx^0) &\triangleq (x_i^t-x_i^0)\int_0^1\!\! \mathrm{d}\alpha  \left.\frac{\partial f}{\partial x_i}\right|_{ \bmx^0 +(\bmx^t-\bmx^0)\alpha}
\end{align}
and $\mathrm{EIG}_i(\bmx^t)\triangleq \!\!
    \int\!\mathrm{d}\bmx^0  P(\bmx^0) \mathrm{IG}_i(\bmx^t\mid \bmx^0)$. 
   Since $\frac{\partial F}{\partial x_i} = \frac{\partial f}{\partial x_i}$, the statement about IG holds. Also, for EIG, the integration w.r.t.~$y^0$ produces $\int \rmd y^0 P(\bmx^0,y^0) = P(\bmx^0)$, yielding $\mathrm{EIG}_i(\bmx^t,y^t) = \mathrm{EIG}_i(\bmx^t)$. 
\end{proof}

\subsubsection{Shapley value}\label{subsubsec:Shapley}

There are a few different versions of SV in the literature~\cite{sundararajan2020many}. Here we adopt the definition of the conditional expectation SV applied to the deviation function:
\begin{align}\label{eq:SV_def_main_text} 
\mathrm{SV}_i(\bm{x}^t,y^t) &= \frac{1}{M}
\sum_{k=0}^{M-1} 
\binom{M-1}{k}^{-1}\!\!\!\!\!
\sum_{\mathcal{S}_i: |\calS_i|=k} \Delta f(\calS_i),
\end{align}
where $\calS_i$ denotes any subset of the variable indices $i \in \{1,\ldots,M\}$ excluding $i$. $|\mathcal{S}_i|$ is the size of $\mathcal{S}_i$. The second summation runs over all possible choices of $\mathcal{S}_i$ under the constraint $|\mathcal{S}_i|=k$ from the first summation. We also define the complement $\bar{\mathcal{S}}_i$, which is the subset of $\{1,\ldots,M\}$ excluding $i$ and $\calS_i$. For example, if $M=12, i=3$ and  $\mathcal{S}_i = \{1,2 \}$, the complement $\bar{\mathcal{S}}_i$ will be $\{4,5,\ldots,12\}$. Corresponding to this division, we rearrange the $M$ variables as $\bmx = ( x_i, \bm{x}_{\mathcal{S}_i}, \bm{x}_{\bar{\mathcal{S}}_i} )$. Finally, the $\Delta f(\calS_i)$ term is defined as the difference between the expected values of $F$ under two different conditions: One is $(x_i,\bm{x}_{\mathcal{S}_i}, y)=(x_i^t, \bm{x}^t_{\mathcal{S}_i},y^t)$ with $\bm{x}_{\bar{\mathcal{S}}_i}$ to be integrated out. The other is $(\bm{x}_{\mathcal{S}_i}, y) = (\bm{x}^t_{\mathcal{S}_i}, y^t)$ with $(x_i,\bm{x}_{\bar{\mathcal{S}}_i})$ to be integrated out. We denote them by $\langle F \mid x_j^t, \bm{x}_{\mathcal{S}_j}^t, y^t\rangle$ and $\langle F \mid \bm{x}^t_{\mathcal{S}_j}, y^t \rangle$, respectively.

The following property holds: 
\begin{theorem}\label{th:SV_deviation_agnositic_main_text}
SV is deviation-agnostic: $\mathrm{SV}_i(\bmx^t,y^t)= \mathrm{SV}_i(\bmx^t)$.
\end{theorem}
\begin{proof}
    Since $F$ is linear in $y$, we can easily see that 
    $
\langle F \mid x_j^t, \bm{x}_{\mathcal{S}_j}^t, y^t\rangle
= \langle f \mid x_j^t, \bm{x}_{\mathcal{S}_j}^t \rangle - y^t
$ and $\langle F \mid \bm{x}^t_{\mathcal{S}_j}, y^t \rangle
= \langle f \mid \bm{x}^t_{\mathcal{S}_j} \rangle - y^t$ hold, which implies $\mathrm{SV}_i(\bmx^t,y^t)= \mathrm{SV}_i(\bmx^t)$.
\end{proof}

\subsection{Relationship between IG, SV, and LIME}\label{subsec:unified_IG_SV_LIME}

The fact that (E)IG, SV, and LIME share the same deviation-agnostic property suggests that they may share a common mathematical structure. In what follows, we show two results showing the interrelationship between them. 

\subsubsection{SV and EIG}
First, let us consider the relationship between SV and EIG. The integral in IG and the combinatorial definition SV are major obstacles in getting deeper insights into what they really represent. This issue can be partially resolved by resorting to power expansion. The following remarkable property holds: 
\begin{theorem}[Equivalence of SV to EIG]
\label{th:SV=EIG_main_text}
a) $\mathrm{SV}_i$ is equivalent to $\mathrm{EIG}_i$ for $\forall i$ up to the second order of the power expansion. b) SV and EIG satisfy exactly the same sum rule:
\begin{align}\label{eq:sum_rule_SV_EIG}
    \sum_{i=1}^M\mathrm{SV}_i(\bmx^t) = \sum_{i=1}^M\mathrm{EIG}_i(\bmx^t) = f(\bmx)-\langle f\rangle,
\end{align}
where $\langle f \rangle \triangleq \int\rmd\bmx\ P(\bmx) f(\bmx)$. 
\end{theorem}
We leave the proof to our companion paper~\cite{Ide2023Arvive_GPA} due to space limitations. While the sum rule b) is known, Theorem~\ref{th:SV=EIG_main_text}~a) is the first result directly establishing the fact that $\forall i, \mathrm{SV}_i \approx \mathrm{EIG}_i$, to the best of our knowledge. In Sec.~\ref{sec:experiments}, we empirically show that indeed SV and EIG systematically give similar attribution scores. 

\subsubsection{LIME and EIG}
Second, let us now consider the relationship between LIME and EIG. LIME, as a local linear surrogate modeling approach, differs from EIG and SV in two regards. First, LIME does not need the true distribution $P(\bmx)$. Instead, it uses a local distribution to populate local samples. Second, LIME is defined as the gradient, not a differential increment. These observations lead us to an interesting question: Is the \textit{derivative of EIG} in the local limit the same as the LIME attribution score? The following theorem answers this question affirmatively: 
\begin{theorem}[LIME and IG] \label{th:LIME=derivative_of_EIG_main_text}
The derivative of IG and EIG is equivalent to LIME:
    \begin{align}
     \!\!\!\mathrm{LIME}^0_i(\bmx^t)\! = \!\lim_{\eta \to 0}\!\!\frac{\partial \mathrm{EIG}_i (\bmx^t) }{\partial x_i}
     = \!\!\!\lim_{\bmx^0 \to \bmx^t}\!\!\frac{\partial \mathrm{IG}_i (\bmx^t \mid \bmx^0) }{\partial x_i},
\end{align}
where the localized Gaussian $P(\bmx^0 )= \calN(\bmx \mid \bmx^t, \eta \sfI_M)$ is used in the definition of EIG. 
\end{theorem}

We leave the proof to our companion paper~\cite{Ide2023Arvive_GPA}. Since EIG, SV, and LIME can be derived from or associated with IG, it is legitimate to say that they are in the \textit{integrated gradient (IG) family}. Since IG is deviation-agnostic, we conclude that the deviation-agnostic property is a common characteristic of the IG family.

\subsubsection{Increment vs. deviation and local vs. global}
\label{subsubsec:implication}

Now let us consider the implications of these results in anomaly attribution. The definition of IG in Eq.~\eqref{eq:IG_x_y_main_text} indicates that IG explains the \textit{increment} of $f(\cdot)$ from the baseline point rather than the deviation, as illustrated in Fig.~\ref{fig:problem_setting_and_deviation_v_increment}. The baseline is arbitrary. Hence, the increment is not directly relevant to the observed anomaly in general. EIG (and thus SV by Theorem~\ref{th:SV=EIG_main_text}) neutralizes this limitation by taking the expectation. However, it results in losing the locality of explanation because it attempts to explain the increment from \textit{any} point in the domain, as suggested in~\cite{kumar2020problems} regarding SV. They are unsuitable for anomaly attribution due to both their deviation-agnostic property and the lack of locality. LIME, on the other hand, maintains the locality by choosing the baseline input in the infinitesimal neighborhood, i.e.,~$\bmx^0 \to \bmx^t$, but it is still deviation-agnostic. 

In general, we need a certain reference point to define anomalousness (cf.~the `reference point' column in Table~\ref{table:comparison chart}). The above observations motivate us to explore a new idea in choosing a reference point. In this regard, the likelihood-based approach first proposed by the present authors~\cite{Ide21AAAI} is quite suggestive. Inspired by~\cite{Ide21AAAI}, we propose a novel generative framework for anomaly attribution, where the notion of normalcy is equated to maximum a posteriori (MAP) estimation, as presented in the next section.

\section{Generative Perturbation Analysis}\label{Sec:MOC}

We have argued that the existing attribution methods are not suitable for anomaly attribution due to their deviation-agnostic property and/or limited built-in mechanism for evaluating the uncertainty of attribution. This section presents the method of generative perturbation analysis (GPA), a novel probabilistic framework for anomaly attribution that addresses these issues.

\subsection{Generative model description}
\label{subsec:seeking_reference_point}

In a typical anomaly detection scenario, samples in the training dataset are assumed to have been collected under normal conditions, and hence, the learned function $y=f(\bmx)$ represents normalcy as well. As discussed in Sec.~\ref{sec:problem_setting}, the canonical measure of anomalousness is the negative log likelihood $-\ln p(y\mid \bmx)$. A low likelihood value signifies anomaly, and vice versa. From a geometric perspective, on the other hand, being an anomaly implies deviating from a certain normal value. We are interested in integrating these two perspectives. 

\subsubsection{Perturbation as explanation}
Suppose we just observed a test sample $(\bmx^t,y^t)$ being anomalous because of a low likelihood value. Given the regression function $y=f(\bmx)$, there are two possible geometric interpretations on the anomalousness (see Figs.~\ref{fig:problem_setting_and_deviation_v_increment}~(b) and~\ref{fig:informative_vs_noninformative}~(a)). One is to start with the input $\bmx = \bmx^t$, and observe the deviation $f(\bmx^t)-y^t$. In some sense, $(\bmx,y)=(\bmx^t,f(\bmx^t))$ is a reference point against which the observed sample $(\bmx^t,y^t)$ is judged. The other is to start with the output $y=y^t$, and move horizontally, looking for a perturbation $\bmdelta$ such that $\bmx = \bmx^t + \bmdelta$ gives the maximum possible fit to the normal model. In this case, the reference point is $(\bmx^t + \bmdelta, y^t)$ and $\bmdelta$ is the deviation measured horizontally. Since $\bmdelta$ is supposed to be zero if the sample is perfectly normal, each component $\delta_1,\ldots,\delta_M$ can be viewed as a value indicative of the responsibility of each input variable. 

\subsubsection{Generative model}
Based on the intuition above, we define a novel probabilistic attribution approach through a data-generating process of observed data. The idea is that we write down a generative process for the observable variables $(\bmx,y)$ as a \textit{parametric model of $\bmdelta$}. Then, \textit{the whole task of anomaly attribution is reduced to a parameter estimation problem}, given an observed test point $(\bmx,y)=(\bmx^t, y^t)$. Specifically, the probabilistic regression model $p(y\mid \bmx)$ is now viewed as a parametric model $p(y\mid \bmx,\bmdelta)$ by setting $\bmx$ to $\bmx + \bmdelta$. With an extra parameter $\lambda$ representing the precision of the regression function and also prior distributions for $\lambda$ and $\bmdelta$, we consider the following generative process:
\begin{gather}\label{eq:GPA_model_gaussian_obs}
     p(y^t\mid \bmx^t, \bmdelta, \lambda) = 
     \left(\frac{\lambda}{2\pi}\right)^{-\frac{1}{2}}
   \exp\left\{ -\frac{\lambda [y^t-f(\bmx^t +\bmdelta)]^2}{2} \right\}\\
     \label{eq:GPA_model_priors_delta}
     p(\bmdelta)= \calN(\bmdelta\mid \bmzero, \eta^{-1}\sfI_M)
     \triangleq (2\pi)^{-\frac{M}{2}} \eta^{-\frac{1}{2}}
   \exp\left\{ -\frac{\eta}{2}\| \bmdelta\|_2^2 \right\}, 
   \\ \label{eq:GPA_model_priors_lambda}
     p(\lambda) = \mathrm{Gam}(\lambda \mid a_0,b_0) \triangleq \frac{b_0^{a_0}}{\Gamma(a_0)}\lambda^{a_0-1}\exp(-b_0\lambda),
\end{gather}
where $\calN(\cdot\mid \cdot,\cdot)$ and $\mathrm{Gam}(\cdot\mid \cdot,\cdot)$ denote the Gaussian and gamma distributions, respectively, and $\eta,a_0,b_0$ are hyperparameters. As mentioned above, $\bmdelta$ plays the role of a model parameter here. Notice that Eq.~\eqref{eq:GPA_model_gaussian_obs} naturally represents the horizontal point-seeking mentioned above. If we point-estimated $\bmdelta$ with Eq.~\eqref{eq:GPA_model_gaussian_obs} alone, we would have the one that achieves $f(\bmx^t +\bmdelta)\approx y^t$. The challenge here is how to find the \textit{distribution} of $\bmdelta$. The prior distribution $p(\bmdelta)$ in Eq.~\eqref{eq:GPA_model_priors_delta} introduces potential variability of $\bmdelta$ to the model. Since $\bmdelta=\bmzero$ represents the normal state, the use of zero-mean Gaussian makes sense. Other zero-mean distributions may work. In fact, we modify this prior a bit later, as discussed in Sec.~\ref{subsec:Solving_MAP}.

\begin{figure}[t]
\begin{center}
\includegraphics[trim={7cm 4.1cm 1.5cm 2.8cm},clip,width=8cm]{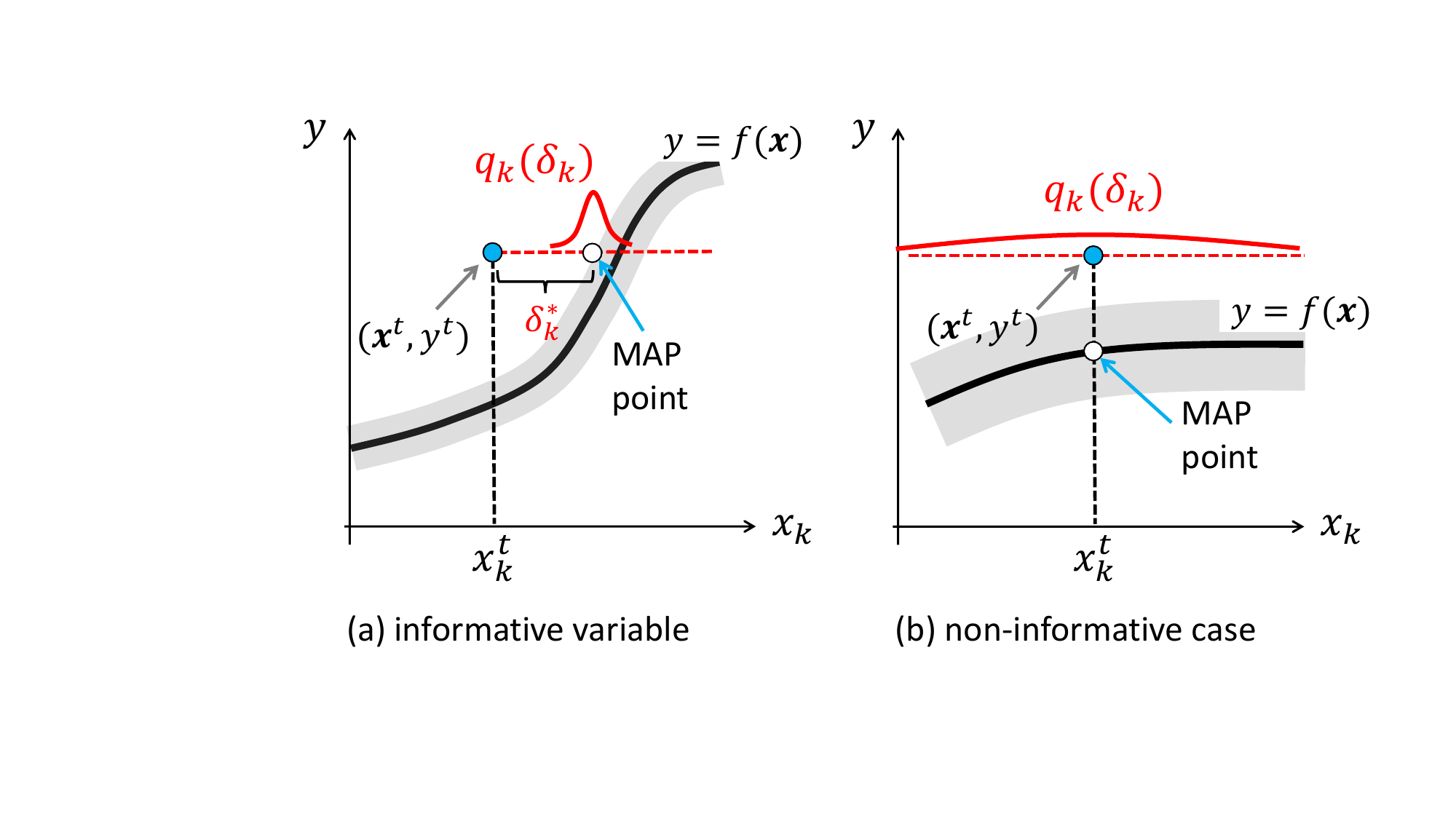}
\end{center}
\vspace{-0.2cm}
\caption{Illustration of the GPA solution. (a)~The MAP estimate $\delta^*_k$ intuitively represents the deviation from the regression surface at the level of $y^t$. (b) If $x_k$ is barely correlated with $y$, GPA tends to give a broad distribution around the MAP point, which is 0 almost surely. }
\label{fig:informative_vs_noninformative}
\end{figure}

The precision parameter $\lambda$ in Eq.~\eqref{eq:GPA_model_gaussian_obs} describes potential noise that may have contaminated the data, as illustrated as the grey band in Fig.~\ref{fig:informative_vs_noninformative}. As the preciseness of the measurements may vary from sample to sample, the use of a single $\lambda$ value can be risky. The prior $p(\lambda)$ takes care of this aspect. As will be seen later, our model uses a mixture of Gaussians with different values of $\lambda$ in some sense, which leads to the $t$-distribution instead of Gaussian for the observation model, adding extra capability of handling heavy noise. 

Finally, we make two remarks about the proposed generative model. First, Bayesian (linear) regression models similar to the above have been considered in the literature, e.g.,~\cite{slack2021reliable}. Our model is fundamentally different from them in that (1) $\bmdelta$ as an explanation is not linear regression coefficients, and (2) we do \textit{not} approximate $f(\cdot)$ as a liner function. As for other Bayesian regression approaches, Moreira et al.~\cite{moreira2021linda} used the Gaussian process in the active learning setting, but not for attribution. 
Second, one might wonder whether the particular choice of a parametric form might lead to the loss of generality. Regarding this question, it is critical to understand that Eq.~\eqref{eq:GPA_model_gaussian_obs} is about the \textit{deviation} or the \textit{error} $f(\bmx^t)-y^t$. Although the variability of $y$ over the entire domain obviously does not follow Gaussian in general, the error is often well-represented by Gaussian or $t$-distribution. This is exactly the same situation Carl Friedrich Gauss faced when he invented Gaussian-based fitting~\cite{brereton2014normal}: Planetary motions do not follow Gaussian, but the error does.

\subsection{Inference approach}
\label{subsec:posterior_inference}

Given the generative model above, the task of probabilistic attribution is now turned to that of finding the posterior distribution of $\bmdelta$. However, there are two major differences from the standard Bayesian inference: 1) $f(\bmx)$ is a black-box function. Exact inference is not possible. Approximating $f(\bmx)$ with a specific functional form, such as the linear function, may not always be possible, either. 2)~Posterior inference generally yields a joint distribution for $\bmdelta$, denoted by $Q(\bmdelta)$. However, this is \textit{not} what we want since it does \textit{not directly explain} the contribution of the \textit{individual} input variables. This section explains how we addressed these challenges.

\subsubsection{Decomposing variable's contributions}

One of the most important ideas of our probabilistic attribution framework is to assume a  \textit{factorized} form of posterior: 
\begin{align}\label{eq:factorized_Q}
\textstyle
   Q(\bmdelta)= Q(\delta_1,\ldots,\delta_M) \approx \prod_{k=1}^M q_k(\delta_k),
\end{align}
so that end-users can directly use $q_k(\delta_k)$ to get insights on the contribution of the $k$-th input variable (see Fig.~\ref{fig:informative_vs_noninformative}). The factorized form~\eqref{eq:factorized_Q} is reminiscent of what is assumed in the variational Bayes (VB) algorithm~\cite{Bishop}, and we can be guided by VB's general solution approach. Specifically, we find the unknown distributions $\{q_k\}$ by minimizing the KL (Kullback–Leibler) divergence between $Q$ and $\prod_k q_k$. The key fact here is that $Q$ is proportional to the complete likelihood by Bayes' rule. Since $\lambda$ is an unobserved intermediate parameter, it can be marginalized. The integration can be performed analytically, yielding the following form of the likelihood:
\begin{align}
    Q&(\bmdelta)\propto p(\bmdelta)\prod_{t=1}^{N_\mathrm{test}}
    \int_0^\infty\rmd\lambda \ p(y^t\mid \bmx^t,\bmdelta,\lambda)p(\lambda)
    \\  \label{eq:t-distributed_Q}
    &\propto p(\bmdelta)\prod_{t=1}^{N_\mathrm{test}}
    \frac{1}{\sqrt{b_0}}\left\{   1 + \frac{[y^t - f(\bmx^t + \bmdelta)]^2}{2b_0}  \right\}^{- (a_0 +\frac{1}{2}) },
\end{align}
where we assumed the collective attribution scenario for generality but note that $N_\mathrm{test}$ can be 1. The marginalization amounts to forming a weighted mixture of Gaussians. The resulting distribution~\eqref{eq:t-distributed_Q} is the $t$-distribution with the degrees of freedom $2a_0$, the mean $f(\bmx^t + \bmdelta)$, and the scale parameter $\sqrt{{b_0}/{a_0}}$, adding extra robustness to the model. 
The objective functional for $\{q_k\}$ is given by
\begin{align}
    \int \left(\prod_{k=1}^M\rmd \delta_k \ q_k(\delta_k)\right) \ln \frac{\prod_{l=1}^M q_l(\delta_l)}{Q(\bmdelta)}
    + \sum_{k=1}^M \gamma_k \! \int\! \rmd \delta_k \ q_k(\delta_k),
\end{align}
where the first term is the KL divergence and the second term is to include the normalization condition with $\gamma_k$ being Lagrange's multiplier. Note that the proportional coefficient in Eq.~\eqref{eq:t-distributed_Q} has no effect here so we do not have to determine it. 

By the calculus of variations w.r.t.~$q_k$, it is straightforward to get the minimizer as
\begin{align} \label{eq:GPA_VB_equation}
    \ln q_k(\delta_k) = \rmc. +  \int \left(\prod_{j \neq k}\rmd \delta_j \ q_j(\delta_j)\right) \ln  Q(\bmdelta),
\end{align}
where $\rmc.$ is a symbol representing an unimportant constant in general. Since both $q_k$ and $q_j$ ($j\neq k$) are unknown, this procedure is iterative in nature. Also, since $q_k$'s are a functional of the black-box function $f(\cdot)$, analytically performing the integration is not possible. Although Monte Carlo techniques can be used in theory, they are not a preferable choice in realistic usage scenarios, where the end-users actively interact with the attribution tool with different test points.

\begin{algorithm}[tb]
\caption{Generative Perturbation Analysis}\label{algo:GPA}
\label{alg:algorithm}
\begin{algorithmic}[1] 
\REQUIRE $f(\bm{x})$, $\mathcal{D}_\mathrm{test}$, parameters $\eta,\nu,\kappa, a_0, \{b(\bmx^t)\}$.
\STATE randomly initialize $\bm{\delta}\approx \bm{0}$.
\REPEAT
\STATE set $\bm{g}=\bm{0}$
\FOR{all $(y^t, \bm{x}^t) \in  \mathcal{D}_\mathrm{test}$ }
    \STATE Compute the local gradient $
\frac{\partial f(\bm{x}^t+\bm{\delta})}{\partial \bm{\delta}}$
    \STATE Update $\bm{g} \leftarrow \bm{g} + \frac{\partial f(\bm{x}^t+\bm{\delta})}{\partial \bm{\delta}} \frac{y^t -f(\bmx^t+\bmdelta)}{2b(\bmx^t) + [y^t -f(\bmx^t+\bmdelta)]^2} $
\ENDFOR
\STATE $\bmg \leftarrow (1 - \kappa\eta)\bm{\delta} + \kappa(2a_0+1) \bm{g}$
\STATE $\bm{\delta} = \sign(\bmg) \max\left\{0, |\bmg| - \eta\nu  \right\}$
\UNTIL convergence
\STATE set $\bmdelta^* = \bmdelta$
\FOR{all $k$}
\STATE $q_k(\delta) = Q(\delta_1^*,\ldots, \delta_{k-1}^*,\delta,\delta_{k+1}^*,\delta,\delta_M^*)$
\STATE $q_k(\cdot) \leftarrow  q_k(\cdot)/\int \rmd \delta' q_k(\delta')$ with Eq.~\eqref{eq:q_k_how_to_normalize}
\ENDFOR
\STATE \textbf{return} $\{ q_k(\cdot) \mid k =1,\ldots,M\}$ and $\bmdelta^*$ \\
\end{algorithmic}
\end{algorithm}

\subsection{Computing attribution score distribution}
\label{subsec:Solving_MAP}

Here, we propose a practical solution to address these 
challenges. For attribution purposes, we do not necessarily need the posterior distribution over the entire domain. What we are interested in is how attribution score is distributed around the most probable value. Hence, we evaluate the expectation in Eq.~\eqref{eq:GPA_VB_equation} through the empirical distribution of $\delta_j$ ($j\neq k$) with a sample at the maximum posteriori (MAP) point. In this approach, the variable-wise posterior is given simply by
\begin{align}\label{eq:VB_mean_field_solution}
    q_k(\delta_k)& \propto Q(\delta_1^*,\ldots, \delta_{k-1}^*,\delta_k,\delta_{k+1}^*,\ldots,\delta_M^*),
\end{align}
where $\bmdelta^*$ is the MAP solution $\bmdelta^* \triangleq \arg\max_{\bmdelta} \ln Q(\bmdelta)$. Since this is a one-dimensional (1D) distribution and we know $\bmdelta$ distributes around zero, the normalization constant can be determined easily. Numerical integration is one approach. Otherwise, one may treat $q_k(\cdot)$ as a discrete distribution on a 1D grid. Specifically, we define 1D grid points $\delta^{[1]},\dots, \delta^{[N_\mathrm{g}]}$ over  $[-\delta_{\max}, \delta_{\max}]$, where $N_\mathrm{g}$ is an arbitrary number of grid points, such as 100, and $\delta_{\max}$ can be, for example, $\delta_{\max}\sim 1.1\max_k |\delta^*_k|$. The distribution $q_k(\cdot)$ on the grid is obtained from its unnormalized version $\tilde{q}_k(\cdot)$ by
\begin{align}\label{eq:q_k_how_to_normalize}
    q_k(\delta^{[i]}) 
    \approx \frac{1}{\sum_{i=1}^{N_\mathrm{g}} \tilde{q}_k(\delta^{[i]})} \tilde{q}_k(\delta^{[i]}), \quad \mbox{for} \ \  i= 1,\ldots, N_\mathrm{g}.
\end{align}

The inference procedure has now become a two-step process: MAP estimation and construction of $\{q_k\}$ with Eqs.~\eqref{eq:VB_mean_field_solution}-\eqref{eq:q_k_how_to_normalize}. The former problem is written as
\begin{gather}\label{eq:GPA_MAP_problem}
\bmdelta^* = \arg\min_{\bmdelta}\left\{ J(\bmdelta) +  \eta\nu \|\bmdelta\|_1 \right\},
\\ \label{eq:J(delta)}
\!\!\! J(\bmdelta)  \triangleq 
 \frac{\eta}{2}\|\bmdelta\|_2^2 + \ln \left\{1 + \frac{[y^t - f(\bmx^t +\bmdelta)]^2}{2b(\bmx^t)}\right\}^{\frac{2a_0+1}{2}}\!\!\!,
\end{gather}
where we have added an extra $\ell_1$ term for better interpretability through sparsity. Here, $\nu$ is the strength of the $\ell_1$ regularization relative to that of $\ell_2$. With this modification, we need to use 
\begin{align}
    p(\bmdelta) \propto \exp\left\{ -\frac{\eta}{2}\| \bmdelta\|_2^2 - \eta\nu \| \bmdelta\|_1\right\}
\end{align}
in Eqs.~\eqref{eq:t-distributed_Q} and~\eqref{eq:VB_mean_field_solution}. We have also included in Eq.~\eqref{eq:J(delta)} potential dependency of $b_0$ on $\bmx^t$ and denoted it as $b(\bmx^t)$. 
Corresponding to $a(\calD_\mathrm{test})$ in Eq.~\eqref{eq:changeScore}, if we wish to find the attribution distribution for a collection of test samples,  $J(\bmdelta)$ should be replaced with
\begin{align}
\!\!\! J(\bmdelta) = 
 \frac{\eta}{2}\|\bmdelta\|_2^2 + \sum_{t=1}^{N_{\mathrm{test}}} \ln \left\{1 + \frac{[y^t - f(\bmx^t +\bmdelta)]^2}{2b(\bmx^t)}\right\}^{\frac{2a_0+1}{2}} \!\!\!\!\!\!\!\!\!\!\!\!.
\end{align}
We call the proposed probabilistic attribution framework the \textbf{generative perturbation analysis} (GPA) hereafter. As illustrated in Fig.~\ref{fig:informative_vs_noninformative}, the GPA distribution $\{q_k\}$ is useful to have in order to evaluate the general informativeness of the input variables. 

\subsubsection{Solving MAP problem}
One of the standard solution approaches to the optimization problem of the type Eq.~\eqref{eq:GPA_MAP_problem} is proximal gradient descent~\cite{parikh2014proximal}, although the unavailability of closed-form expression of the gradient of $f(\cdot)$ makes the procedure a bit complicated. If a numerical estimation method for $\nabla f(\bmx^t + \bmdelta)$ is available, Eq.~\eqref{eq:GPA_MAP_problem} can be reduced to an iterative lasso regression problem:
\begin{align}
        \!\!\!\bmdelta \leftarrow \arg \min_{\bmdelta}\left\{
        \frac{1}{2\kappa}\| \bmdelta\! - \!\bmdelta' \!+ \!\kappa \nabla J(\bmdelta') \|_2^2 \!+\! \eta \nu \| \bmdelta\|_1\!
        \right\},\!
\end{align}
where $\kappa$ is a constant corresponding to the learning rate and $\bmdelta'$ is the solution of the previous iteration round. By setting the subgradient zero, the solution of this problem is readily obtained as
\begin{align}
    \delta_i = \sign(g_i) \max\left\{0, |g_i| - \eta\nu  \right\},
\end{align}
where we defined $\bmg \triangleq \bmdelta' -\kappa \nabla J(\bmdelta')$. Upon convergence, we set $\bmdelta^*=\bmdelta$. See lines 2-11 in Algorithm~\ref{algo:GPA}.

\subsubsection{Algorithm summary}
Algorithm~\ref{algo:GPA} summarizes the entire algorithm of GPA. Whenever possible, it is recommended to standardize~$\bmx$ somehow so that it distributes around zero with unit variance for each variable. For standardized data, the $\ell_2$ strength $\eta$ can be a value of $\mathrm{O}(1)$, such as 0.1, which can also be a reasonable starting point for $\kappa$. The $\ell_1$ strength should be in the range $0 < \nu\leq 1$. We fixed $\nu=0.5$ in our experiment. As $2a_0$ has the interpretation of degrees of freedom, one reasonable starting point is $2a_0 \sim N_\mathrm{test} +1$. As described in Appendix~\ref{app:hyper_parameter_tuning}, $b(\bmx^t)$ can be chosen as a constant $b_0\sim a_0 \sigma_{yf}^2/c_b$, where $\sigma_{yf}^2$ is an estimate of the variance of $y-f(\bmx)$, or the maximizer of the marginalized likelihood, and $c_b$ is the number of virtual samples, which can be $\mathrm{O}(10)$. As summarized in Table~\ref{table:datasets}, we used $c_b=1$ or 10, and also $2a_0=11$ in our experiments to simulate the variability of realistic cases. 

It is easy to see that the complexity of the algorithm is $O(M N_\mathrm{test})$ per iteration round. Note that $N_\mathrm{test}=1$ is the most common choice (i.e., local explanation) and the algorithm does not use any training data. Hence, typical scalability analysis about the data set size is irrelevant. The total computational time depends almost entirely on very low-level implementation details, such as how efficient the numerical gradient estimation routine is, how the black-box model $f(\cdot)$ is implemented, and to what extent the Python code is vectorized. Their detailed analysis is beyond the scope of the paper.

\begin{table}[tb]
\caption{Summary of the datasets and parameters used. }
\label{table:datasets}
\vspace{-0.3cm}
\footnotesize
\begin{tabular}{lllll | lll}
\hline \hline
                & $N_{\mathrm{train}}$ & $N_{\mathrm{test}}$ & $M$ & $f(\bmx)$  & $\kappa$ & $c_\mathrm{b}$ & $\eta$  \\ \hline
$\mathtt{2Dsinusoidal}$   & $\infty$          & 1                   & 2   & analytic  & - & - & -   \\
$\mathtt{Diabetes}$        & 442                  & 1                   & 10  & DNN  & 0.08 & 10 & 0.4 \\
$\mathtt{Boston}$        & 506                  & 1                   & 13  & RF  & 0.08 & 10& 0.1    \\
$\mathtt{California}$      & $20\,640$            & 3                   & 8 & GBT & $0.1/N_\mathrm{test}$ & 1 & $0.5 N_\mathrm{test}$  \\
\hline 
\end{tabular}
\end{table}

\section{Experiments}\label{sec:experiments}

This section presents empirical evaluation of the proposed anomaly attribution framework\footnote{Python implementation is available at \url{https://github.com/Idesan/gpa}.}. The goals of this evaluation are to 1) provide a clear picture of what deviation-sensitivity of an attribution method buys us; 2) demonstrate GPA's unique capability of providing the probability distribution of attribution scores; 3) quantitatively analyze the consistency and inconsistency among different attribution methods.

\subsection{Datasets and baselines}\label{subsec:baselines}

Based on the datasets summarized in Table~\ref{table:datasets}, we compared GPA with seven baselines: Six non-distributional attribution methods,  LIME~(Sec.~\ref{subsubsec:LIME}), (E)IG~(Sec.~\ref{subsubsec:IG}), SV~(Sec.~\ref{subsubsec:Shapley}), LC~\cite{Ide21AAAI}, and the $Z$-score (e.g.,~\cite{Chandola09AnomalySurvey}), as well as one distributional method,  BayLIME~\cite{zhao2021baylime}. For anomaly attribution, LIME, SV, IG, and EIG are applied to the deviation $f(\bmx) - y $ rather than $f(\bmx)$. The $Z$-score is a standard univariate outlier detection metric in the unsupervised setting, and is defined as 
$ Z_i \triangleq (x_i^t-m_i)/\sigma_i$  
for the $i$-th variable, where $m_i,\sigma_i$ are the mean and the standard deviation of $x_i$, respectively. In SV, we used the same sampling scheme as that proposed in~\cite{vstrumbelj2014explaining} with the number of configurations limited to 100. In IG and EIG, we used the trapezoidal rule with 100 equally-spaced intervals to perform the integration w.r.t.~$\alpha$. For IG, EIG, LC, and GPA, we used the same gradient estimation algorithm described in Appendix~\ref{Appendix:gradient_etimation}. 

To compute SV, EIG, and the $Z$-score, we used the empirical distribution of the training data to approximate $P(\bmx)$. Note that this is \textit{actually not possible to do} in our doubly black-box setting. We are including SV, EIG, and the $Z$-score here for comparison purposes nonetheless.

\begin{figure}[tb]
\centering
\includegraphics[trim={4cm 0.1cm 0cm 0.5cm},clip,width=4.1cm]{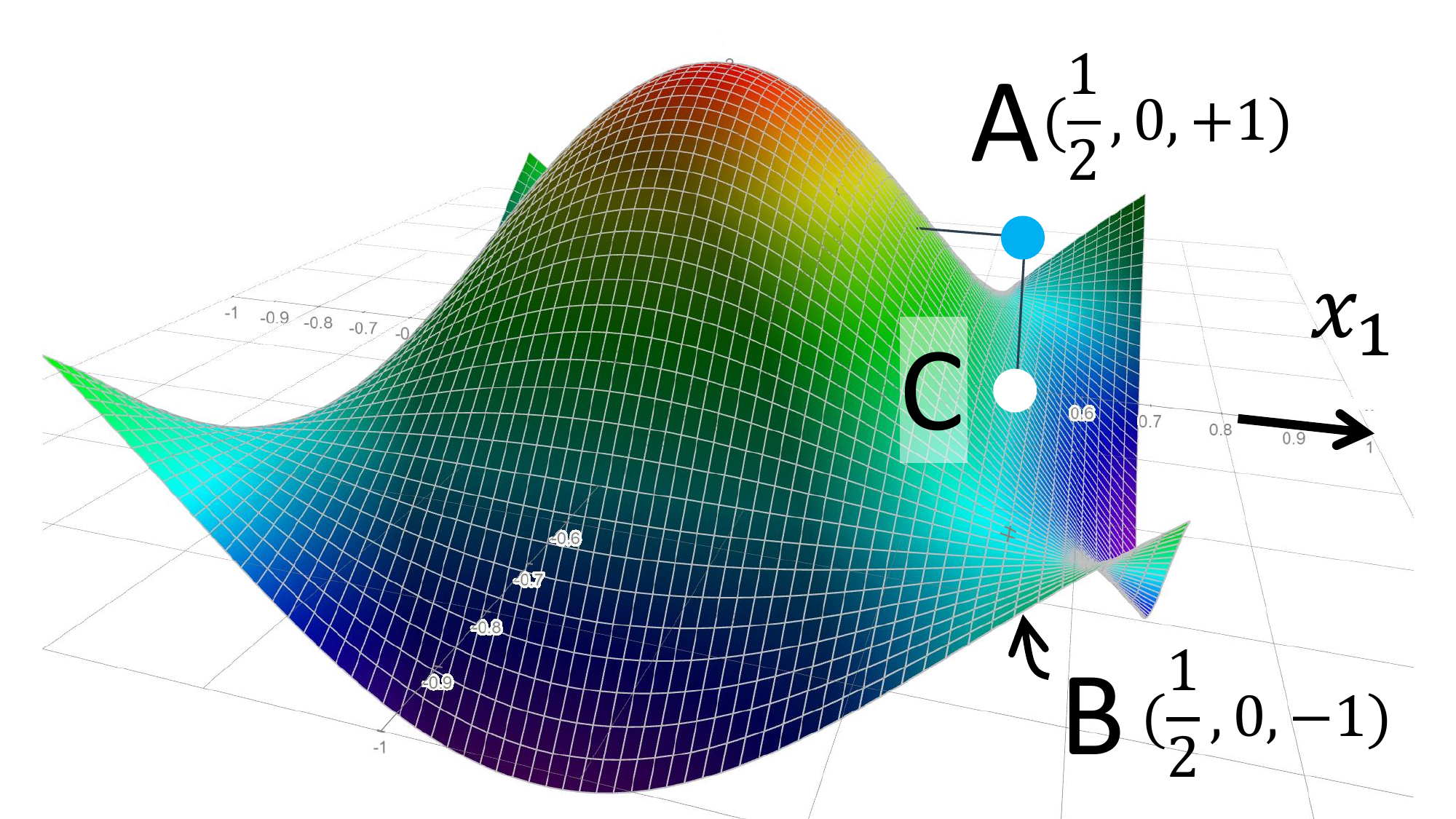}
\hspace{-0.5mm}
\includegraphics[trim={0.2cm 0.cm 0cm 0.4cm},clip,width=4.2cm]{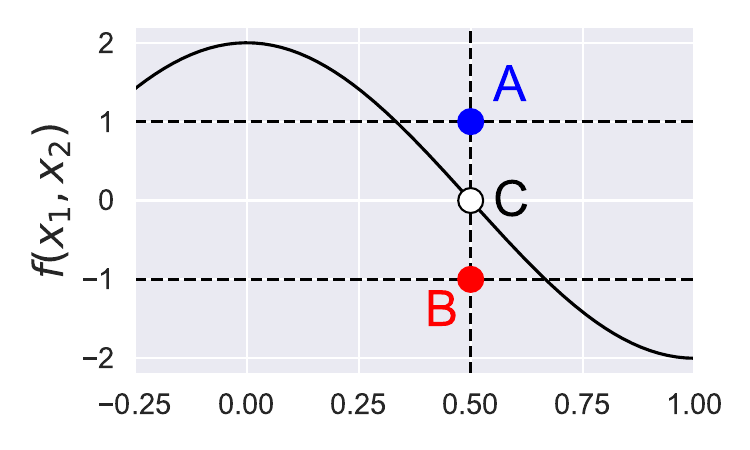}
\vspace{-2mm}
\caption{$\mathtt{2Dsinusoidal}$: Surface plot and the $x_2=0$ slice. The points A, B, and C are at $y^t=1$, $-1$, and $0$, respectively, while they are at the same $\bmx^t =(1/2,0)$. }
\label{fig:2D_sinusoidal}
\end{figure}
\begin{figure}[bt]
\centering
\includegraphics[trim={0.5cm 0.cm 0cm 0.cm},clip,width=0.48\textwidth]{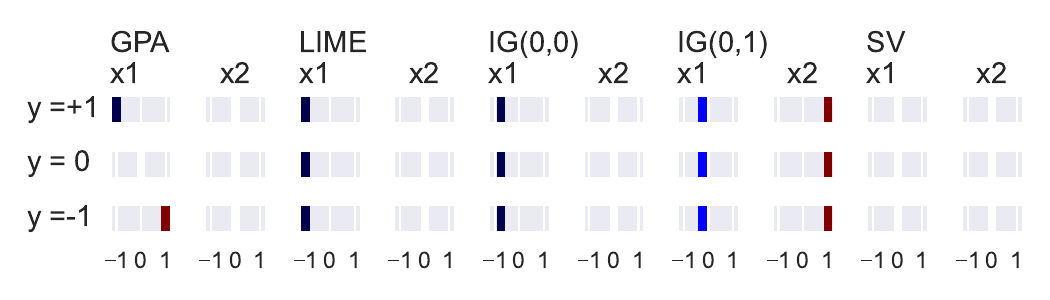}
\vspace{-0.6cm}
\caption{$\mathtt{2Dsinusoidal}$: Comparison of normalized attribution scores at three test points (A, B and C in Fig.~\ref{fig:2D_sinusoidal}).  }
\label{fig:2D_sinusoidal_scores}
\end{figure}

\subsection{Deviation-sensitivity}
\label{sec:Deviation-sensitiveness}

\subsubsection{2Dsinusoidal}

The first empirical evaluation uses a synthetic dataset named $\mathtt{2Dsinusoidal}$, which is a newly proposed attribution benchmark model, defined by a 2-variate sinusoidal function
\begin{align}\label{eq:2d_sinusoidal}
    f(\bmx) &= 2 \cos (\pi x_1)\cos(\pi x_2).
\end{align}
One remarkable feature of this model is that it is possible to \textit{calculate closed-form attribution scores}. See Appendix~\ref{app:2D_sinusoidal_closed_form} for the details. 

%

Suppose we have a test point at $\bmx^t=(1/2,0)$ with three different $y^t$ values, A ($y^t=1$), B ($y^t=-1$), and C ($y^t=0$), as illustrated in Fig.~\ref{fig:2D_sinusoidal}. In this case, SV and LIME attribution scores are $(0,0)$ and $(-2\pi,0)$, respectively. IG's scores are $(-2, 0)$ and $(-2/3,8/3)$ if we choose $\bmx^0 =(0,0)$ and $(0,1)$, respectively. These do not depend on $y^t$ due to the deviation-agnostic property, in contrast to GPA, which gives $\delta_1^*=(1/\pi)\arccos\left(y^t/2\right) -x^t_1$ and $\delta_2^*=0$. 

Figure~\ref{fig:2D_sinusoidal_scores} visualizes the attribution scores with what we call the `\textit{litmus plot},' where larger values get darker colors (0 gets white) and negative/positive values get blue/red colors. Due to space limitations, we omitted LC, which results in the same solution as that of GPA, and the $Z$-score. For GPA, the scores are normalized by dividing by $\max_k |\delta_k|$ for each test point. Similar normalization was done for the baselines with the convention $\frac{0}{0}=0$. 

In this example, A and B are outliers due to a shift in the $x_1$ direction, while C is normal in terms of deviation. Hence, an ideal attribution would be that $x_1$ \textit{alone} gets a strong signal \textit{only} for A and B. GPA precisely reproduces this, but all the baselines do not: They gave the same score for A, B, and C, as a consequence of the deviation-agnostic property. The figure also shows that IG's scores sensitively depend on the choice of $\bmx^0$, making IG trickier to use for the end-users. SV always satisfies $\mathrm{SV}_1=\mathrm{SV}_2$ regardless of $\bmx^t$ in this case, and does not provide any clue for input attribution. This is a manifestation of the loss of locality in SV discussed in Sec.~\ref{subsubsec:implication}.

\subsubsection{Diabetes}

\begin{figure}[tb]
\centering
\includegraphics[trim={0.cm 0.3cm 0cm 0cm},clip,width=4.2cm]{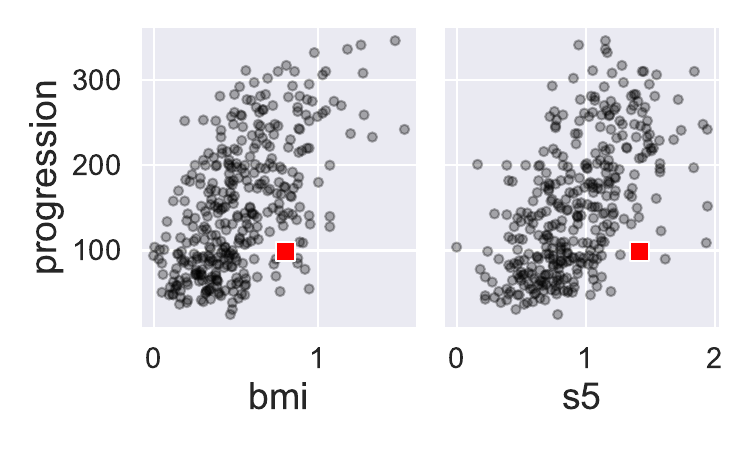}
\includegraphics[trim={0.cm 0.3cm 0cm 0cm},clip,width=4.2cm]{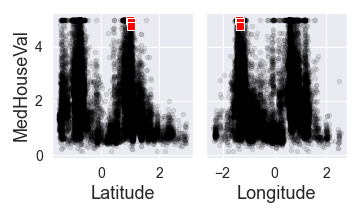}
\vspace{-0.4cm}
\caption{Scatter plot of selected input variables vs.~$y$. Left: $\mathtt{Diabetes}$. Right: $\mathtt{California Housing}$. The red squares highlight the detected top outliers.}
\label{fig:diabetes_scatter_bmi_s5}
\end{figure}

\begin{figure}[tb]
\centering
\includegraphics[trim={0.cm 0.3cm 0cm 0cm},clip,width=8cm]{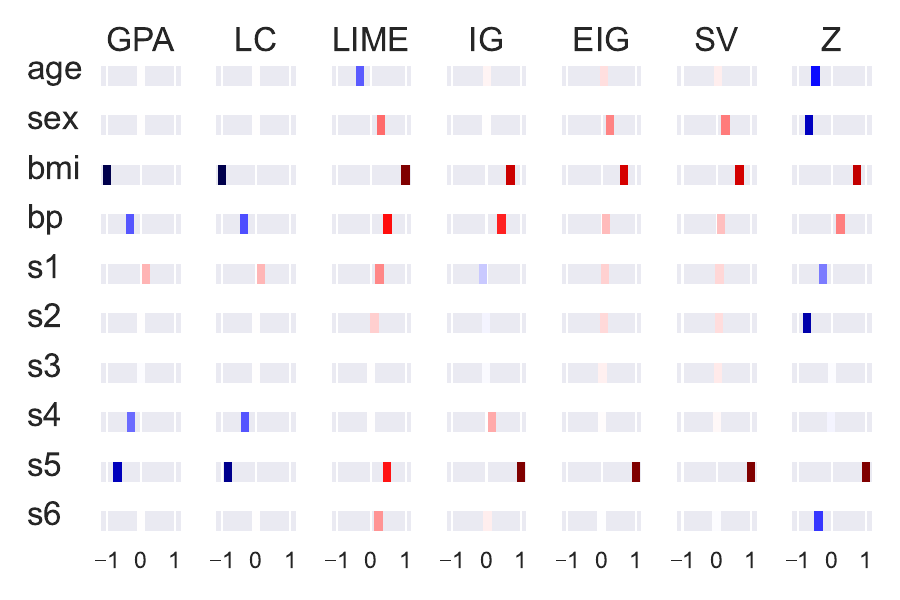}
\vspace{-0.3cm}
\caption{$\mathtt{Diabetes}$: Comparison of normalized attribution scores in the litmus plot for the top outlier detected.}
\label{fig:diabetes_MAP_comaparison}
\end{figure}

To test the deviation sensitivity of GPA on real-world data, we used $\mathtt{Diabetes}$~\cite{efron2004least}, which has a real-valued target variable (`progression') and $M=10$ predictors including the body-mass index (`bmi'). For this dataset, we held out 20\% of the samples and trained a deep neural network (DNN) on the rest as the black-box model $f(\cdot)$. We identified the top outlier using Eq.~\eqref{eq:outlier_score}, which is highlighted in Fig~\ref{fig:diabetes_scatter_bmi_s5}.  

Figure~\ref{fig:diabetes_MAP_comaparison} compares attribution scores for the top outlier. We set $\bmx^0=\bmzero$ for IG.  All the attribution methods identify `bmi' and `s5' as the top contributors. For both, GPA and LC get a large negative score since a smaller value is more typical for such a low $y^t$ value, as shown in the scatter plot in Fig.~\ref{fig:diabetes_scatter_bmi_s5}. GPA gave $\delta_{\mathrm{bmi}}=-0.81$ and $\delta_{\mathrm{s5}}=-0.55$. Note that these values have actual meaning rather than just the magnitude of responsibility: A big negative in $\delta_{\mathrm{bmi}}$ means that the BMI is too high for such a low $y$ level. In other words, they would have looked normal if they were a little skinnier. 
Explainability of this kind is particularly useful in practice as the score provides \textit{actionable insights} about how the status quo could be changed for the better. The alternative methods do not have such ability. LIME is positive for bmi because the slope is positive at the $\bmx^t$, regardless of the $y^t$ value. Similarly, IG, EIG, and SV gave positive values for bmi because $f(\bmx^t)$ is higher than the mean of $y$, regardless of the specific value of $y^t$.

\begin{figure}[tb]
\centering
\includegraphics[trim={0.cm 0.3cm 0cm 0cm},clip,height=6cm]{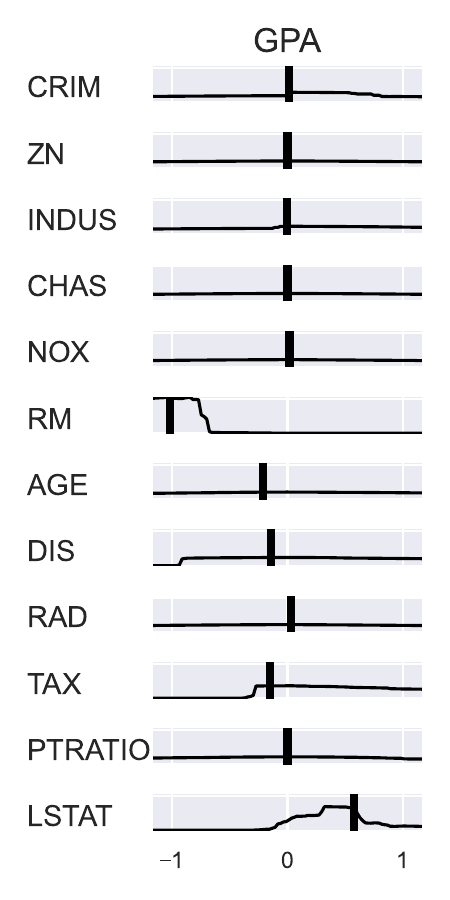}
\includegraphics[trim={0.cm 0.3cm 0cm 0cm},clip,height=6cm]{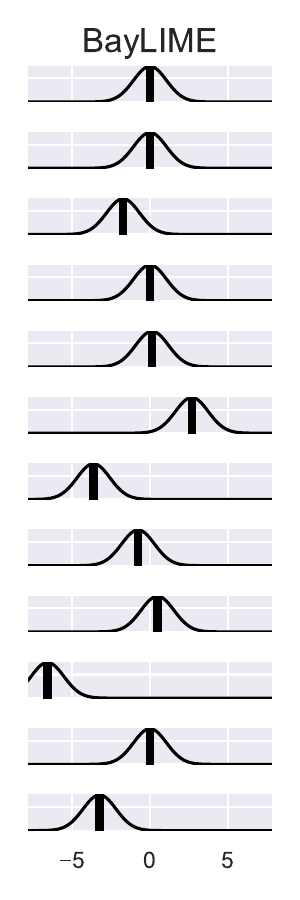}
\includegraphics[trim={0.cm 0.3cm 0cm 0cm},clip,height=6cm]{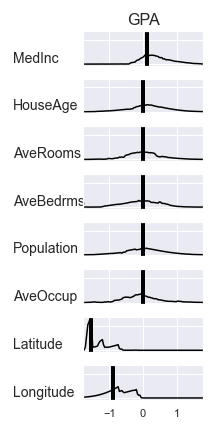}
\vspace{-0.3cm}
\caption{Estimated score distribution. Left: $\mathtt{Boston Housing}$. Right: $\mathtt{California Housing}$ for collective attribution.}
\label{fig:Boston_GPA_distribution}
\end{figure}

\subsection{Distribution analysis}
\label{sec:Distribution_analysis}

\subsubsection{Comparison with BayLIME}
Now let us discuss how GPA provides useful insights into the uncertainty of the attribution score. To the best of our knowledge, BayLIME~\cite{zhao2021baylime,slack2021reliable} is the only method in the literature applicable to our setting. 

We used the $\mathtt{Boston Housing}$ data~\cite{belsley2005regression}, where the task is to predict $y$, the median home price (`MEDV') of the districts in Boston, with $\bmx$, the input vector of size $M=13$ including such features as the percentage of the lower status of the population (`LSTAT') and the average number of rooms (`RM')\footnote{We excluded a variable named `B' from attribution for ethical concerns~\cite{sklearn113_Boston_Housing}.}. For this dataset, we held out 20\% of the samples as $\calD_\mathrm{test}$ and trained the random forest (RF)~\cite{ESL2} on the rest as the black-box model $f(\cdot)$. We identified the top outlier using Eq.~\eqref{eq:outlier_score} and computed the score distribution for the top outlier. 

The estimated distributions are shown in Fig.~\ref{fig:Boston_GPA_distribution}. As clearly seen from the figure, BayLIME gives the same curve for all the variables apart from the mean locations. In fact, the variance is given as a constant $1/(\eta+\lambda N_\mathrm{s})$, where $N_\mathrm{s}$ is the number of virtual samples generated for estimating the regression coefficients and is 10 in our case (See Appendix~\ref{Appendix:ByLIME}). In contrast, GPA provides variable-specific distributions. As illustrated in Fig.~\ref{fig:informative_vs_noninformative}, less informative variables tend to produce a flatter distribution in GPA. In this case, we immediately see that RM and LSTAT are two dominating variables. Such an insight is not obtainable from BayLIME. It is interesting to see that the score distributions given by GPA tend to be piece-wise constant, reflecting the fact that RF is a collection of decision stumps.

\subsubsection{Collective attribution}

To show the unique capability of GPA for collective attribution, we used the $\mathtt{California Housing}$~\cite{pace1997sparse} dataset. The task is to predict the median house value of small geographical segments using predictor variables such the longitude and latitude. We held out 20\% of the samples and trained gradient boosted trees (GBT)~\cite{friedman2002stochastic} on the rest. In this case, we identified \textit{three} top outliers as shown in Fig.~\ref{fig:diabetes_scatter_bmi_s5}. The question is whether those outliers have common characteristics in their outlier-ness. 

Figure~\ref{fig:Boston_GPA_distribution} shows the computed distributions, where we omitted BayLIME due to its triviality. Very interestingly, `Latitude' has a very sharp peak at a negative value. This indicates that the very high `MedHouseVal' in Fig.~\ref{fig:diabetes_scatter_bmi_s5} stands out in that latitude and they would look more common if they existed in a southern location.

\subsection{Consistency analysis}
\label{sec:Consistency_analysis}

We have compared GPA with seven alternative methods in a rather qualitative fashion so far. One important question in practice is how those methods are consistent or inconsistent among them overall. To answer this question, we identified five top outliers in the three real-world datasets ($\mathtt{Boston Housing}$, $\mathtt{California Housing}$, $\mathtt{Diabetes}$), and computed how their attribution scores are consistent with those of GPA in terms of four metrics: Kendall's $\tau$, Spearman's $\rho$, the sign match ratio (SMR), and the hit ratio at 25\% (h25). See Appendix~\ref{app:metrics} for the detail. 

\begin{table}[tb]
\centering
\caption{Result of consistency analysis. The mean and the standard deviation are shown in each cell, where 1 represents the highest consistency with GPA's MAP value.}
\label{table:comprehensive_comparision_consistency}
\footnotesize
\addtolength{\tabcolsep}{-1.2pt}
\begin{tabular}{lllllll}
\hline \hline
                          &            & LC & LIME & IG  & SV & $Z$-score \\
\hline
\multirow{4}{*}{\texttt{Bos.}}   & $\tau$ & $0.70 \pm 0.14$ & $0.25 \pm 0.30$ & $0.61 \pm 0.11$ & $0.43 \pm 0.10$ & $0.17 \pm 0.30$ \\
                          & $\rho$ & $0.83 \pm 0.09$ & $0.32 \pm 0.38$ & $0.74 \pm 0.08$ & $0.57 \pm 0.14$ & $0.24 \pm 0.35$ \\
                          & SMR & $0.92 \pm 0.11$ & $0.71 \pm 0.11$ & $0.65 \pm 0.12$ & $0.69 \pm 0.14$ & $0.62 \pm 0.17$ \\
                          & h25 & $0.80 \pm 0.18$ & $0.27 \pm 0.28$ & $0.73 \pm 0.28$ & $0.67 \pm 0.00$ & $0.20 \pm 0.30$ \\
\hline
\multirow{4}{*}{\texttt{Cal.}}   & $\tau$ & $0.82 \pm 0.15$ & $0.67 \pm 0.13$ & $0.64 \pm 0.07$ & $0.73 \pm 0.10$ & $0.04 \pm 0.20$ \\
                          &$\rho$ & $0.91 \pm 0.11$ & $0.76 \pm 0.11$ & $0.79 \pm 0.06$ & $0.83 \pm 0.10$ & $0.07 \pm 0.27$ \\
                          & SMR  & $0.97 \pm 0.06$ & $0.95 \pm 0.11$ & $0.68 \pm 0.07$ & $0.68 \pm 0.11$ & $0.70 \pm 0.14$ \\
                          & h25 & $0.80 \pm 0.27$ & $0.90 \pm 0.22$ & $1.00 \pm 0.00$ & $1.00 \pm 0.00$ & $0.30 \pm 0.27$ \\
\hline
\multirow{4}{*}{\texttt{Dia.}} & $\tau$ & $0.94 \pm 0.06$ & $0.31 \pm 0.19$ & $0.72 \pm 0.08$ & $0.58 \pm 0.10$ & $0.15 \pm 0.15$ \\
                          &$\rho$ & $0.98 \pm 0.03$ & $0.41 \pm 0.21$ & $0.88 \pm 0.04$ & $0.75 \pm 0.10$ & $0.22 \pm 0.20$ \\
                          & SMR & $1.00 \pm 0.00$ & $0.62 \pm 0.11$ & $0.38 \pm 0.08$ & $0.62 \pm 0.22$ & $0.60 \pm 0.10$ \\
                          & h25    & $1.00 \pm 0.00$ & $0.60 \pm 0.22$ & $0.90 \pm 0.22$ & $0.80 \pm 0.27$ & $0.30 \pm 0.45$ \\
\hline
\end{tabular}
\vspace{-1mm}
\end{table}

The result is summarized in Table~\ref{table:comprehensive_comparision_consistency}. We omitted EIG because of Theorem~\ref{th:SV=EIG_main_text}. LC achieves very high consistency with GPA, although it lacks a built-in mechanism for UQ. This is understandable since it can be viewed as a point-estimation version of GPA in some sense. As expected, h25 generally has high scores, apart from the $Z$-score. This suggests that those attribution methods are a useful tool for selecting important features. Even in the other metrics, including the SMR, they produce reasonably consistent attributions in some cases. However, in some 20-30\% of cases they are not necessarily consistent, which is a natural consequence of the fact that GPA is deviation-sensitive but the others are not. 

\subsection{Practical utility of the GPA framework} 

We remark further on the practical utility of the proposed  framework, using the \texttt{BostonHousing} data as an example. 
Recall that the top detected outlier has two variables with dominating attribution scores. 
Depending on one's role, different insights may be obtained from this analysis: 
From the \textbf{end user}'s perspective, the outlier in Fig.~\ref{fig:Boston_GPA_distribution} may point to a bargain since this house (district) has unusually more rooms and much fewer low-income neighbors than expected for the price range; 
For a \textbf{modeler} who is interested in debugging the model, the two dominating attribution scores may hint that the model may be failing to capture the relationship between the housing price and the variables RM and LSTAT, prompting the modeler to revise (e.g. contextualize) how these variables are defined.
While the attribution scores may not decisively pinpoint the exact interpretation, the rich and accurate information given by GPA provides valuable clues in either usage scenario. 

\section{Conclusions}

We have proposed GPA, a novel generative approach to probabilistic attribution of black-box regression models. The key idea is to reduce the attribution task to a statistical parameter estimation problem. This can be done by viewing the perturbation $\bmdelta$ as a model parameter of the generative process for the observed variables $(\bmx,y)$, where the posterior distribution gives the distribution of the attribution score. We proposed a variational inference algorithm to obtain variable-wise distributions. 

We have also shown that the existing input attribution methods, namely integrated gradient IG), local linear surrogate modeling (LIME), and Shapley values (SV), are inherently deviation-agnostic and, thus, are not designed to be a viable solution for {\em anomaly attribution}. Unlike these methods, GPA is capable of providing directly interpretable insights in a deviation-sensitive and uncertainty-aware manner.


\bibliography{ide_et_al}
\bibliographystyle{ACM-Reference-Format}

\clearpage
\appendix
\section*{Appendix}

\renewcommand{\thesection}{\Alph{section}}
\renewcommand{\theequation}{\Alph{section}.\arabic{equation}}
\renewcommand{\thefigure}{\Alph{section}.\arabic{figure}}
\renewcommand{\thetable}{\Alph{section}.\arabic{table}}
\renewcommand{\thetheorem}{\Alph{section}.\arabic{theorem}}

\setcounter{equation}{0}
\setcounter{section}{0}
\setcounter{figure}{0}
\setcounter{table}{0}
\setcounter{theorem}{0}

\section{Closed-form solutions for 2-variate sinusoidal model}
\label{app:2D_sinusoidal_closed_form}

This section lists analytic expressions of a few attribution methods on the \texttt{2Dsinusoidal} model
$
    f(x_1,x_2) = 2 \cos (\pi x_1)\cos (\pi x_2).
$

\subsection{LIME}

Since LIME score is an estimator of the gradient w.r.t.~the input variables $x_1,x_2$ in the limit of $\nu \to 0_+$, we have
\begin{align}
    \mathrm{LIME}^0(\bmx^t,y^t) = \begin{pmatrix}
    -2\pi \sin (\pi x_1)\cos (\pi x_2)\\
    -2\pi \cos (\pi x_1)\sin (\pi x_2)
    \end{pmatrix}
\end{align}
for $\forall (\bmx^t,y^t)$, which obviously does not depend on $y^t$. If we choose $\bmx^t=(1/2,0)^\top$, then 
$\mathrm{LIME}^0 = (-2\pi, 0)^\top$.

\subsection{GPA}

In GPA, the $J$ function is given by
\begin{align*}
    J(\bmdelta)  \triangleq 
 \frac{\eta}{2}\|\bmdelta\|_2^2 + \ln \left\{1 + \frac{[y^t - f(\bmx^t +\bmdelta)]^2}{2b(\bmx^t)}\right\}^{\frac{2a_0+1}{2}}
\end{align*}
With $\Delta^t_{\bmdelta} \triangleq y^t - f(\bmx^t +\bmdelta)$, the gradient is computed as
\begin{align*}
    \frac{\partial J}{\partial \bmdelta} 
    &= \eta \bmdelta - 
    \frac{(2a_0 + 1 )\Delta^t_{\bmdelta}}{2b(\bmx^t) + (\Delta^t_{\bmdelta})^2} \frac{\partial f(\bmx+\bmdelta)}{\partial \bmdelta},
\end{align*}
where
\begin{align*}
   \frac{\partial f(\bmx+\bmdelta)}{\partial \bmdelta} =     \begin{pmatrix}
    - 2\pi \sin (\pi (x_1^t+\delta_1))\cos (\pi (x_2^t+\delta_2))\\
    - 2\pi \cos (\pi (x_1^t+\delta_1))\sin (\pi (x_2^t+\delta_2))
    \end{pmatrix}.
\end{align*}
Let us assume $\nu \to 0_+$ and $-2 < y^t < 2$. If we assume $x_2^t = 0$, then   
$
\delta_2^* = 0
$ 
should hold as long as $\bmdelta$ is initialized as $\bmdelta \approx \bmzero$ and $\eta\nu>0$ regardless of the sign of $\Delta^t_{\bmdelta}$. Given this partial solution, the condition of optimality for $\delta_1$ is given by
\begin{align*}
    \eta\delta_1 + \left(2a_0 + 1 \right)
    \frac{2\pi \Delta^t_{\bmdelta}  \sin  (\pi(x^t_1+\delta_1))}{2b(\bmx^t) + (\Delta^t_{\bmdelta})^2}  = 0.
\end{align*}
If $x^t_1 > 0$ and $\eta \to 0_+$, this equation yields a condition
$
    y^t -2 \cos (\pi( x_1+\delta_1)) \approx 0,
$ 
leading to the solution
\begin{align}
    \delta^*_1 = \frac{1}{\pi}\arccos\frac{y^t}{2} -x^t_1.
\end{align}
If we further choose $x^t_1=1/2$ (i.e., $\bmx^t =(1/2,0)$ again), we have 
$\bmdelta^*=(-\frac{1}{6},0)^\top$, $(0,0)^\top$, $(\frac{1}{6},0)^\top$ for $y^t=1,0$, $-1$, respectively.

\subsection{LC}

In LC, the $J$ function in our notation is given by
\begin{align}
    J(\bmdelta) =\frac{1}{2}\eta\| \bmdelta\|_2^2
    + \frac{1}{2}\lambda[y^t - f(\bmx^t + \bmdelta)]^2.
\end{align}
With $\Delta^t_{\bmdelta} \triangleq y^t - f(\bmx^t +\bmdelta)$, the gradient is computed as 
\begin{align*}
    \frac{\partial J}{\partial \bmdelta} 
    &= \eta\bmdelta -\lambda \Delta^t_{\bmdelta}
    \begin{pmatrix}
    - 2\pi \sin (\pi (x_1^t+\delta_1))\cos (\pi (x_2^t+\delta_2))\\
    - 2\pi \cos (\pi (x_1^t+\delta_1))\sin (\pi (x_2^t+\delta_2))
    \end{pmatrix}.
\end{align*}
Let us assume $\nu \to 0_+$ and $-2 < y^t < 2$. If we assume $x^t_2 = 0$, again, $\delta_2^* = 0$ should hold as long as $\bmdelta$ is initialized as $\bmdelta \approx \bmzero$ and $\eta\nu>0$ regardless of the sign of $\Delta^t_{\bmdelta}$. Given this partial solution, the condition of optimality for $\delta_1$ is written as
\begin{align}
    \eta\delta_1 + \lambda[y^t -2 \cos (\pi( x_1+\delta_1))] 2\pi \sin  (\pi(x^t_1+\delta_1)) = 0.
\end{align}
If $x^t_1 > 0$ and $\eta\to 0_+$, we have a condition 
$
    y^t -2 \cos (\pi( x_1+\delta_1)) \approx 0
$, 
which is the same as the MAP equation of GPA. Hence, LC gets the same attribution score as GPA's MAP value in this particular case.

\subsection{Integrated Gradient}

The \texttt{2Dsinusoidal} model allows calculating IG analytically for any $(\bmx^t,\bmx^0)$ based on the definition~\eqref{eq:IG_x_y_main_text} as 
\begin{align}
    \mathrm{IG}_i(\bmx^t,y^t \mid \bmx^0,y^0)=d_i\left[ G^t-G^0 -(-1)^i(H^t-H^0)\right]
\end{align}
with $i$ being 1 or 2 and 
\begin{align}
    G^k \triangleq \frac{\cos \pi(x_1^k +x_2^k) }{d_1 +d_2}, \quad 
H^k \triangleq
\frac{\cos \pi(x_1^k - x_2^k) }{d_1 -d_2},
\end{align}
where $d_1 \triangleq x^t_1 - x^0_1,\ d_2 \triangleq x^t_2 - x^0_2$ and $k$ is either $t$ or $0$. Using elementary trigonometric formulas, one can verify the sum rule $\mathrm{IG}_1 + \mathrm{IG}_2 = f(\bmx^t) - f(\bmx^0)$. For $\bmx^t  =(1/2,0)^\top$, the IG values are 
\begin{align*}\textstyle
    \mathrm{IG}(\bmx^t \mid  (0,0)^\top) = (-2,0)^\top, \quad 
     \mathrm{IG}(\bmx^t \mid (0,1)^\top)=
     ( -\frac{2}{3},  \frac{8}{3})^\top,
\end{align*}
where we have omitted redundant $y^t,y^0$ from the arguments.

\subsection{Shapley Value}

The expected Shapley value depends on the true distribution $P(\bmx)$. If $P(\bmx)$ is the uniform distribution over $[-m,m]$ with $m$ being an integer, the expectation of $f$ is zero in \texttt{2Dsinusoidal}. The same applies to the conditional distributions. As a result, we have
\begin{align}
    \mathrm{SV}(\bmx^t) = \frac{1}{2}
(   f(\bmx^t), f(\bmx^t) )^\top
\end{align}
for $\forall \bmx^t$ under the assumed uniform distribution.

\section{Attribution score distribution with Bayesian LIME}
\label{Appendix:ByLIME}

Equation~(7) of BayLIME's paper~\cite{zhao2021baylime} provides the posterior distribution of the regression coefficients. In our notation, the posterior covariance is given by 
\begin{align}
    \bm{\Sigma}^{-1} = \eta \sfI_M + \lambda \sum_{n=1}^{N_s} \bmxi_n \bmxi_n^\top,
\end{align}
where $\bmxi_n$ is the $n$-th sample generated from $\calN(\bmxi \mid \bmzero, \sfI_M)$, according to the authors. The paragraph after their Eq.~(11) says that $\sum_{n=1}^{N_s} \bmxi_n \bmxi_n^\top \approx N_s\sfI_M$ holds. Hence, $\bm{\Sigma}$ can be computed as
\begin{align}
    \bm{\Sigma} =\left\{ \eta \sfI_M + \lambda N_s \sfI_M\right\}^{-1}
    = (\eta +\lambda N_s )^{-1}\sfI_M
\end{align}
and the posterior distribution of the attribution score $\bmbeta$ is given by $Q^{\mathrm{BayLIME}} \triangleq \calN(\bmbeta \mid \bmbeta^{\mathrm{BayLIME}}, \bm{\Sigma})$, where $\bmbeta^{\mathrm{BayLIME}}$ is the posterior mean. Since $\bm{\Sigma}$ is diagonal, $\beta_i$s are statistically independent. The distribution of the attribution score of the $k$-th variable is given by
\begin{align}
q_k^{\mathrm{BayLIME}}(\beta_k) =
    \calN(\beta_k \mid \beta^{\mathrm{BayLIME}}_k, (\eta + \lambda N_s)^{-1}).
\end{align}
This is a one-dimensional distribution with the same variance for all the $k$s. Since the model evaluates the \textit{variability of the generated samples based on an assumed distribution}, the variance does not have any explicit dependency on the black-box function $f(\cdot)$.

\section{Estimating the gradient of black-box function}
\label{Appendix:gradient_etimation}

To find the MAP solution for GPA, we need to numerically estimate the gradient of the black-box function $f(\cdot)$.  To handle the potential non-differentiability of $f$, we define the gradient as the local mean of the slope function $[f(\bmx_\delta + h\bme_i) - f(\bmx_\delta)]/h$, where $\bmx_\delta\triangleq\bmx^t+\bmdelta$, $h$ is a small random perturbation, and $\bme_i$ is a unit vector which takes 1 in the $i$-th entry and 0 otherwise. The local mean can be estimated by numerically evaluating
\begin{align}\label{eq:gradient_estimation_as_mean_slope}
     \frac{\partial f(\bmx_\delta)}{\partial \delta_i} 
    &=
    \int^\infty_{-\infty} \rmd h \ p(h) \frac{f(\bmx_\delta + h\bme_i) - f(\bmx_\delta)}{h},
\end{align}
where $p(h)$ is a local distribution for $h$ around $\bmx_\delta$. One reasonable choice is $p(h )=\calN(h \mid 0,\eta_1^2)$ with $\eta_1$ being the standard deviation of the perturbations. For numerical stability, we used $\eta_1=1$ in our experiments, where the input variables have been standardized. The number of Monte Carlo samples was set to 10, which was confirmed to provide sufficient convergence in our experiments.

\section{Parameter tuning approach}\label{app:hyper_parameter_tuning}

GPA's distribution can be used for verifying whether the computed MAP value has reached a satisfactory local maximum. In our experiments, we started with a default set of parameters: $\kappa =0.1/N_\mathrm{test}$, $\eta=0.1 N_\mathrm{test}$, and $c_b=10$. If any of the GPA distributions looked inconsistent with the MAP value, we gradually decreased $c_b$ down to 1 and increased $\eta$ up to 1. We kept $\nu$ fixed at $0.5$, which turned out to achieve a sparsity level comparable to that of LIME. 

We discuss how to initialize $a_0,b_0$ in the gamma prior below.

\subsection{Gamma hyper-parameters: shape}
\label{app:Gamma_hyperparameter_a}

Since $2a_0$ has the interpretation of the degree of freedom of the $t$-distribution, it makes sense to use 
\begin{align}
    a_0 = (\tilde{N}+1)/2.
\end{align}
Here, $\tilde{N}$ denotes the sample size and can be equated to $N_\mathrm{test}$. We have added 1 so $a_0=1$ when $N_{\mathrm{test}}=1$. Otherwise, it can be interpreted as the virtual sample size, which can be a measure of the confidence level. Since UQ generally boils down to an ill-posed task that estimates uncertainty somehow when many samples are not available, $\tilde{N}$ can be viewed as a controllable parameter to simulate what would be seen when there were abundant samples. In such a case, $\tilde{N}$ could be a value like $1 \leq \tilde{N} \lesssim 10$.

\subsection{Gamma hyper-parameters: rate}
\label{app:Gamma_hyperparameter_b}

Given $a_0$, the other parameter $b_0$ can be estimated by maximizing the log likelihood. For $\Delta_n \triangleq y^{(n)} - f(\bmx^{(n)})$, we solve
\begin{align*}
    &\max_b\sum_{n=1;n \neq t}^{N_\mathrm{test}} w_n(\bmx^t)\left\{\rmc. -\frac{1}{2}\ln b - \left(a_0 + \frac{1}{2}\right)
    \ln \frac{1}{b}\left( b + \frac{\Delta_n^2}{2} \right) \right\}
    \\
    &=
    \max_b\sum_{n=1;n \neq t}^{N_\mathrm{test}} w_n(\bmx^t)\left\{\rmc. + a \ln b - \left(a_0 + \frac{1}{2}\right)
    \ln \left( b + \frac{\Delta_n^2}{2} \right) \right\}
\end{align*}
to obtain an iterative formula
\begin{align}\label{eq:b_iteration_PRLC}
    \frac{1}{b(\bmx^t)} \leftarrow \frac{2a_0 + 1}{a_0} \sum_{n=1;n \neq t}^{N_\mathrm{test}}   \frac{\tilde{w}_n(\bmx^t)}{2 b(\bmx^t) + [y^{(n)} - f(\bmx^{(n)})]^2},
\end{align}
where $\tilde{w}_n \triangleq \frac{w_n}{\sum_m w_m}$. For the kernel function, we can use, e.g., 
\begin{align}\label{eq:Gaussian_kernel}
    w_n(\bmx^t) = w_0 + \exp\left(-\frac{\|\bmx^{(n)} -\bmx^t \|^2}{2 \eta_0^2}\right).
\end{align}

We need an initial estimate for $b_0$. One reasonable choice is obtained by replacing $ [y^{(n)} - f(\bmx^{(n)})]^2$ with its average $\sigma^2_{yf}$, yielding
\begin{align}\label{eq:b_0_initiatlization}
    b_0 \approx a_0 \sigma_{yf}^2, \quad \mbox{where}\quad \sigma^2_{yf}  \triangleq \frac{1}{N_{\mathrm{test}}}\sum_{t=1}^{N_\mathrm{test}} [y^t - f(\bmx^t)]^2.
\end{align}
Recall that the derived $t$-distribution has the scale parameter $\sqrt{b_0/a_0}$. As the scale parameter corresponds to the standard deviation, we see that the above relationship $b_0/a_0 \sim \sigma^2_{yf}$ is consistent with it.

Equation~\eqref{eq:b_0_initiatlization} can be also used as a constant approximation for $b(\bmx^t)$. However, for evaluating the probability density function of $\bmdelta$, it tends to give a bit too large value. This is understandable because if, e.g., $N_\mathrm{test}=1$, a majority of the probability mass is from the prior, giving a dull peak around zero. To reproduce a realistic distribution, we need to `simulate' the situation where there are a reasonable number of test samples. This can be done by choosing a smaller $b_0/a_0$ because the precision (the reciprocal of the variance) linearly increases as a function of the sample size in Bayesian estimation. Hence, when estimating the distribution in GPA, we can include a correction factor $c_b$ as
\begin{align}
    b_0 \sim a_0\sigma^2_{yf}/c_b.
\end{align}
Intuitively, $c_b$ is interpreted as the number of virtual parameters. Typically, $c_b \sim 10$ gives a reasonable distribution but it should be viewed as a free parameter that can be tuned according to each use-case.

\section{Comparing attribution scores}\label{app:metrics}

We computed the following four metrics to evaluate the consistency among different attribution methods. The first and second metrics are Kendall's $\tau$ and Spearman's $\rho$, calculated for two \textit{absolute} attribution score vectors. They take a value of 1 if the orders are the same regardless of their values. The third metric is what we call the sign match ratio (SMR), which takes on 1 when all the signs are consistent between corresponding vector elements. When comparing an attribution score vector $\bmu$ against a reference score vector $\bmr$, SMR is defined as
\begin{align}
    (\mbox{SMR}) &\triangleq 1 - \frac{1}{M}\sum_{i=1}^M\mathbb{I}\left(\sign(r_i)\sign(u_i) = -1\right),
\end{align}
where $\mathbb{I}(\cdot)$ is the indicator function that takes on 1 when the argument  is true, 0 otherwise. We define $\sign(0)=0$ in this case. Note that this favors sparse attribution scores: If $\bmr =\bmzero$, then the score is always 1 regardless of $\bmu$. Finally, the fourth metric is what we call hit25, which gives 1 when the top 25\% of the absolute entries perfectly match between $\bmr$ and $\bmu$, and 0 if none of the top 25\% members of $\bmr$ is included in that of $\bmu$. As hit25 depends on neither the sign nor the rank, it quantifies simply the match of top contributors.  

\end{document}